\documentclass{article} 
\usepackage{iclr2026_conference,times}

\usepackage{subcaption}
\usepackage{graphicx}
\usepackage{url}
\usepackage{caption}
\usepackage{enumitem}
\usepackage{booktabs}
\usepackage{multirow}
\usepackage{makecell}
\usepackage{hyperref}
\usepackage{colortbl}  
\usepackage[most]{tcolorbox}
\usepackage{placeins}
\usepackage{xcolor}
\usepackage{tabularx}
\usepackage{float}
\usepackage{CJKutf8}
\usepackage[capitalize]{cleveref}

\title{Long-range Modeling and Processing of\\ Multimodal Event Sequences}

\author{Jichu Li$^{1}$\thanks{Equal contribution.}\enspace, Yilun Zhong$^{1}$\footnotemark[1]\enspace, Zhiting Li$^{1}$, Feng Zhou$^{1,2}$\thanks{Corresponding authors.}\enspace, Quyu Kong$^{3}$\footnotemark[2]\\
$^{1}$Center for Applied Statistics and School of Statistics, Renmin University of China\\
$^{2}$Beijing Advanced Innovation Center for Future Blockchain and Privacy Computing\\
$^{3}$Independent Researcher\\
\texttt{\{lijichu52,kongquyu\}@gmail.com}, \texttt{feng.zhou@ruc.edu.cn}
}

\iclrfinalcopy 
\begin{document}

\maketitle
\thispagestyle{fancy} 
\fancyhead[L]{Published as a conference paper at ICLR 2026} 

\begin{abstract}
Temporal point processes (TPPs) have emerged as powerful tools for modeling asynchronous event sequences. While recent advances have extended TPPs to handle textual information, existing approaches are limited in their ability to generate rich, multimodal content and reason about event dynamics. A key challenge is that incorporating multimodal data dramatically increases sequence length, hindering the ability of attention-based models to generate coherent, long-form textual descriptions that require long-range understanding. In this paper, we propose a novel framework that extends LLM-based TPPs to the visual modality, positioning text generation as a core capability alongside time and type prediction. Our approach addresses the long-context problem through an adaptive sequence compression mechanism based on temporal similarity, which reduces sequence length while preserving essential patterns. 
We employ a two-stage paradigm of pre-training on compressed sequences followed by supervised fine-tuning for downstream tasks. Extensive experiments, including on the challenging DanmakuTPP-QA benchmark, demonstrate that our method outperforms state-of-the-art baselines in both predictive accuracy and the quality of its generated textual analyses. Code is publicly
available at \href{https://github.com/JichuLi/MM-TPP}{https://github.com/JichuLi/MM-TPP}.
\end{abstract}

\section{Introduction}

Temporal point processes (TPPs)~\citep{daley2003introduction,daley2007introduction} are statistical models designed for analyzing asynchronous event sequences in continuous time. They have found widespread applications across diverse domains including neuroscience~\citep{zhou2021efficient}, finance~\citep{bacry2015hawkes}, crime analysis~\citep{mohler2013modeling}, and epidemiology~\citep{rizoiu2018sir}. While traditional statistical TPPs can model event timing and dependencies, they often lack the expressiveness needed to capture complex real-world patterns. Recent advances in deep learning have led to the development of deep TPPs~\citep{du2016recurrent,mei2017neural,simiao2020transformer,meng2024interpretable}, which demonstrate superior modeling capabilities. Building on the success of large language models (LLMs), Language-TPP~\citep{kong2025language} further extended TPPs to incorporate textual information, showing the potential of integrating natural language processing with temporal modeling. 

However, real-world event sequences are increasingly becoming multimodal, extending beyond just temporal and textual information. For instance, in video streaming platforms, user comment streams (known as Danmaku) contain not only timestamps and textual content but also associated visual frames from the video~\citep{jiang2025danmakutppbench}. Similarly, traffic monitoring systems capture accident reports that include temporal information, textual descriptions, audio recordings, and visual evidence from surveillance cameras. Current TPP models, including Language-TPP, remain largely confined to limited modalities and lack the capability to effectively model and generate rich multimodal content such as images alongside text and temporal information. 


Extending deep TPPs to be fully generative in multimodal settings presents two primary challenges.
\textbf{First, existing TPP models lack a unified framework for multimodal generation.} While Language-TPP has progressed in incorporating text, it cannot handle visual content or generate text that is conditioned on it. Different modalities require distinct encoding strategies~\citep{lahat2015multimodal}, and a truly generative framework must not only encode this heterogeneous data but also produce contextually aware content—such as generating a textual description for a future event that correctly references its associated visual information. To address this, we propose a unified framework that extends Language-TPP's template-based approach to incorporate visual information, enabling comprehensive multimodal event modeling and generation. 

\textbf{Second, incorporating multimodal data dramatically increases sequence length, hindering the ability to generate coherent, long-form text that depends on long-range dependencies.}
This inflation occurs because non-textual modalities are themselves tokenized into sequences; for example, a single image is commonly decomposed into hundreds of patch-based tokens~\citep{dosovitskiy2021imageworth16x16words}. In event sequences where each point is associated with such data, the total length $N$ can become exceptionally large. This makes the $\mathcal{O}(N^2)$ complexity of the Transformer's~\citep{vaswani2017attention} self-attention mechanism a critical bottleneck. This prevents the model from accessing the full history needed to generate a comprehensive analytical report or answer summary questions. To mitigate this, we propose an adaptive compression mechanism that leverages temporal similarity to reduce sequence length while preserving the dynamics required for effective long-range text generation. 

In summary, to address the above challenges, we propose \textbf{Multimodal Temporal Point Processes (MM-TPP)}, a unified framework that tackles these issues through the following contributions: 

\textbf{(1)} We propose MM-TPP, an extension of Language-TPP to a multimodal framework that predicts event time and type while generating rich text conditioned on visual, textual, and temporal inputs. 

\textbf{(2)} We introduce an adaptive compression mechanism based on temporal similarity that reduces sequence length, making the generation of long-form, context-aware text computationally feasible. 

\textbf{(3)} We construct TAXI-PRO, a new multimodal TPP dataset by augmenting the classic NYC Taxi dataset with visual map patches and textual descriptions, providing a valuable benchmark. 

\textbf{(4)} Comprehensive experiments on two multimodal TPP datasets, including the DanmakuTPP-QA benchmark, demonstrate the superiority of MM-TPP in predictive accuracy compared to SOTA baselines. Furthermore, qualitative analysis reveals that our model generates significantly more insightful and coherent textual analyses, showcasing its superior reasoning capabilities.
\vspace{-0.45em}

\section{Related Works}
\paragraph{Deep TPPs.}
Deep learning-based TPPs were first introduced by \citet{du2016recurrent} using an RNN-based framework. Follow-up works improved this approach, including \citet{mei2017neural, yang2018recurrent, omi2019fully, soen2021unipoint, chen2024non}. Recurrent models offer efficient inference, with constant-time prediction and linear training cost, but suffer from limited parallelism and poor modeling of long-range dependencies. 
To address these issues, Transformer-based TPPs emerged, starting with \citet{simiao2020transformer} and \citet{zhang2020self}, and later extended by works such as \citet{zhu2021deep, zhou2022neural, yang2022transformer, meng2024interpretable, panosdecomposable}. These models enable parallel training and better capture long-range dependencies via self-attention, but require higher computational cost—both time and memory scale as $\mathcal{O}(N^2)$. 
In contrast to these approaches, MM-TPP not only handles long-range dependencies efficiently but also extends TPP modeling to generate and reason over rich multimodal content. 

\paragraph{Covariate TPPs.}
Covariate TPPs enhance traditional models by incorporating contextual information to better explain event dynamics and improve predictions. Early works focused on structured covariates—e.g., \citet{meyer2012space} used demographic data for epidemic modeling, \citet{xiao2017modeling} proposed a dual-LSTM for combining covariates and event history, and \citet{adelfio2021including, gajardo2021point} applied similar ideas to earthquake and COVID-19 data. 
In practice, many covariates are unstructured, such as text, images, or audio, making their integration into TPPs challenging. Some studies tackled this by using word frequency features \citep{zhu2022spatiotemporal} or BERT embeddings \citep{zhang2023integration}. More recent efforts leverage LLMs to jointly model temporal and textual information, as in \citet{liu2024tpp, kong2025language}. 
MM-TPP builds on this direction by incorporating multiple types of unstructured covariates—specifically, images and text—into a unified TPP framework, and further enables multimodal generation for future events. 


\begin{figure*}[t]
    \centering
    \begin{subfigure}[t]{0.494\textwidth}
        \centering
        \includegraphics[width=\linewidth, height=4cm, page=1]{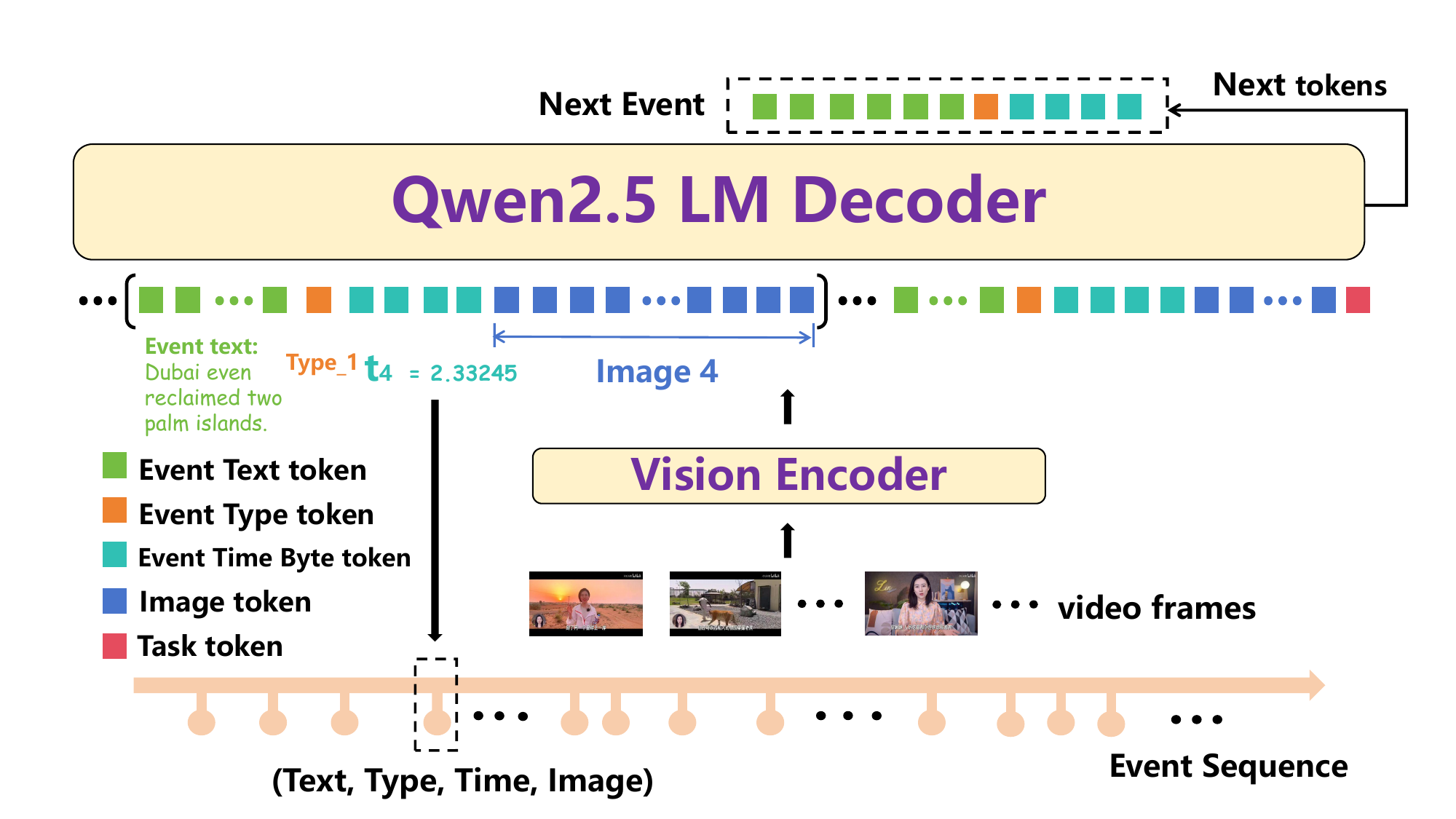}
        \caption{}
        \label{subfig:mainfig_1}
    \end{subfigure}
    \hfill
    \begin{subfigure}[t]{0.494\textwidth}
        \centering
        \includegraphics[width=\linewidth, height=4cm, page=2]{iclr2026/main_fig.pdf}
        \caption{}
        \label{subfig:mainfig_2}
    \end{subfigure}
    \caption{Overview of the MM-TPP framework. (a) Model Architecture: The framework is built upon Qwen2.5-VL, where multimodal events—consisting of time, type, text, and image information—are tokenized into unified token sequences. Visual content is encoded by the vision encoder and fused with other modalities via the language model decoder to perform autoregressive next-event prediction. 
    (b) Adaptive Long Sequence Compression: Events with similar time intervals (\( |\tau_i - \tau_{i-1}| < \Delta \)) are compressed into a special token. This strategy enables efficient processing of long event sequences while preserving temporal structure and key event information.
    }
    \label{fig:mainfig}
\end{figure*}

\section{Preliminaries}

This section introduces the foundational concepts and background knowledge for our proposed model, including TPPs and LLM-based TPPs modeling. 

\subsection{Temporal Point Processes}
TPPs are probabilistic models designed for sequences of discrete events occurring in continuous time. A typical realization of a TPP is denoted by $\mathcal{S} = \{(t_i, e_i)\}_{i=1}^{N}$, where each $(t_i, e_i)$ represents an event of type $e_i \in \mathcal{E} = \{1, \dots, E\}$ that occurs at time $t_i \in (0, T]$, satisfying $0 < t_1 < \dots < t_N \le T$. Here, $N$ denotes the random number of events observed in the time window $(0, T]$. 
There are multiple equivalent formulations to characterize TPPs. One common approach is to model the conditional distribution of future events given the past. Let $\mathcal{H}_{t}$ denote the event history up to and including time $t$. 
We define $p(t_{i+1}, e_{i+1} | \mathcal{H}_{t_i})$ as the conditional density function for the next event occurring at time $t_{i+1}$ with type $e_{i+1}$, given the past history $\mathcal{H}_{t_i}$.
Under this formulation, the probability of observing a sequence $\mathcal{S}$ can be factorized as follows: 
\begin{equation*}
    p(\mathcal{S}) = \prod_{i=1}^{N} p(t_i, e_i \mid \mathcal{H}_{t_{i-1}})\cdot \left(1 - P(T \mid \mathcal{H}_{t_N})\right), 
\end{equation*}
where $1 - P(T \mid \mathcal{H}_{t_N})$ represents the probability that no events occur in $(t_N, T]$, $P(T \mid \mathcal{H}_{t_N}) = \int_{t_N}^{T} p(t \mid \mathcal{H}_{t_N})\,\mathrm{d}t$ is the cumulative density of the next event time after $t_N$, and $p(t \mid \mathcal{H}_{t_i}) = \sum_{e=1}^{E} p(t, e \mid \mathcal{H}_{t_i})$. 
It is standard to decompose the conditional joint density as: 
\[
p(t, e \mid \mathcal{H}_{t_i}) = p(t \mid \mathcal{H}_{t_i}) \cdot p(e \mid \mathcal{H}_{t_i}, t). 
\]
Therefore, based on the above formulations, the log-likelihood of a TPP can be derived as follows: 
\begin{equation}
\begin{aligned}
\log p(\mathcal{S}) 
&= \sum_{i=1}^N  \log p(t_i | \mathcal{H}_{t_{i-1}}) + \log p(e_i | \mathcal{H}_{t_{i-1}}, t_i) + \log \left(1 - P(T \mid \mathcal{H}_{t_N})\right). 
\end{aligned}
\label{eq:CDF_likelihood}
\end{equation}

\subsection{Language-TPP}
Recent advances in LLMs have enabled new approaches to TPP modeling. Early studies adapted pre-trained LLMs to temporal event data using customized tokenization and fine-tuning~\citep{liu2024tpp, xue2023prompt, shi2024language}. Following this direction, \citet{kong2025language} proposed Language-TPP, a framework for modeling event sequences of the form ${(t_i, e_i, m_i)}_{i=1}^N$, where each event includes a timestamp, type, and textual description. The model captures dependencies across the sequence and jointly predicts future event time, type, and description. 

To achieve this, Language-TPP adopts a decoder-only causal Transformer and introduces several components. Each event is tokenized by applying a standard language tokenizer to $e_i$ and $m_i$, while $t_i$ is encoded using a byte-level strategy with 256 special tokens, allowing each 32-bit timestamp to be represented by four tokens. The resulting tokens are organized into a structured event template, where each event is enclosed by \texttt{<|start\_of\_event|>} and \texttt{<|end\_of\_event|>} tokens, and each component is prefixed by a modality-specific marker (e.g., \texttt{<|time\_prefix|>} for timestamps). The model then performs autoregressive generation until an end-of-sequence token is produced, from which the timestamp, type, and description of future events can be decoded. 
Language-TPP is trained in two stages. First, event sequences are converted into token sequences and the backbone LLM is pre-trained using standard next-token prediction. In the second stage, prompt-response pairs are constructed for each subtask (time prediction, type classification, and description generation), and fine-tuning is performed on the response portion using next-token loss.

While Language-TPP effectively incorporates textual information into TPP modeling, it is limited to a single modality and faces scalability issues when handling long multimodal sequences. Our proposed MM-TPP extends this framework by incorporating both visual and textual modalities, and introduces a temporal similarity-based compression mechanism to address the sequence length explosion in multimodal settings. 

\section{Methodology}
\label{methodology}
To model multimodal event sequences with rich visual and textual information under long-context constraints, we propose MM-TPP, an extension of Language-TPP. It jointly models temporal, categorical, textual, and visual data, and introduces a compression mechanism to enhance long-range dependency handling. 

\subsection{Model Framework}
\cref{subfig:mainfig_1} provides an overview of the framework of MM-TPP. 
Our model is built upon Qwen2.5-VL~\citep{bai2025qwen2}, a SOTA pre-trained multimodal LLM. The entire framework follows a sequence-to-sequence paradigm, taking a multimodal event history sequence as input and autoregressively predicting various attributes of future events. 
Extending the annotations from~\citet{kong2025language}, given a multimodal event sequence ${(t_i, e_i, m_i, v_i)}_{i=1}^N$, where $t_i$ denotes the event timestamp, $e_i$ represents the event type, $m_i$ is the textual description, and $v_i$ is the associated visual description, the goal is to model the dependencies within such a sequence and predict the time, type, textual content of future events.

\subsection{Multimodal Event Sequence Tokenization}
\label{template}

To handle multimodal event data with language models, we design a tokenization scheme that converts each event tuple $(t_i, e_i, m_i, v_i)$ into a unified token sequence suitable for autoregressive modeling. Temporal information is encoded following the design of Language-TPP, where each 32-bit time interval is decomposed into four bytes based on its memory layout. These bytes are mapped to 256 special tokens, \texttt{<|byte\_0|>} to \texttt{<|byte\_255|>}, allowing for compact and precise representation of time intervals. Categorical information, such as event types, is handled using dedicated special tokens (e.g., \texttt{<|type\_0|>} to \texttt{<|type\_5|>} for six event categories), enabling unified and interpretable modeling of discrete labels.

For textual information, event descriptions (e.g., user comments) are tokenized using the built-in tokenizer of Qwen2.5-VL, which converts raw text into standard language tokens that can be directly processed by the model. Visual information is integrated differently: instead of converting image pixels into tokens, we insert a placeholder token \texttt{<|image\_pad|>} at the appropriate position in the sequence. During model execution, the corresponding image is passed through Qwen2.5-VL’s vision encoder to produce visual embeddings, which are then aligned with the placeholder token. This design allows for deep fusion of visual, textual, and temporal features while keeping the token sequence compact and language-model-compatible.

To organize these multimodal components, we define a structured event template. Each event is enclosed by \texttt{<|start\_of\_event|>} and \texttt{<|end\_of\_event|>} tokens. Within the event, each modality is marked by a prefix token: \texttt{<|time\_start|>} for time, \texttt{<|type\_start|>} for type, \texttt{<|text\_start|>} for text, and \texttt{<|vision\_start|>} for image. This structured template provides a consistent and interpretable format for encoding complex multimodal events. A full example of the uncompressed event format is shown in \cref{ap:temp}.

\subsection{Adaptive Long Sequence Compression}

The inclusion of text and image tokens in multimodal event sequences significantly increases input length, posing challenges for Transformer-based LLMs due to high computational cost and limited context windows. To mitigate this, we introduce an adaptive compression strategy based on temporal similarity between events. In many real-world settings—such as video comment streams—events arrive in bursts or follow periodic rhythms, resulting in similar inter-event intervals. MM-TPP leverages this observation by compressing sequences of temporally similar events, as shown in \cref{subfig:mainfig_2}.

During input construction, we define a temporal similarity threshold \( \Delta \) and compare the interval \( \tau_i = t_i - t_{i-1} \) of the current event with that of the previous one \( \tau_{i-1} \). If the difference between the two is smaller than the threshold, i.e., \( |\tau_i - \tau_{i-1}| < \Delta \), event \( i \) is deemed temporally similar to event \( i-1 \). In this case, instead of encoding it using the full multimodal template, it is replaced with a single special token \texttt{<|similar\_event|>}, thereby reducing the number of tokens. Details on choosing an appropriate value for \( \Delta \) are discussed in \cref{ap:similar}. Otherwise, if the time intervals differ significantly, event \( i \) is represented in full using the multimodal template defined in \cref{template}. For a concrete step-by-step illustration of this sequential compression mechanism applied to consecutive events, please refer to \cref{app:compression_mechanism}.

This strategy enables compact representation of dense event clusters—potentially requiring hundreds of tokens—using only a few \texttt{<|similar\_event|>} tokens, while preserving key events with distinctive timing patterns. As a result, MM-TPP can utilize the fixed context window (e.g., 4096 tokens) more effectively, allowing it to model longer event histories and capture long-range dependencies with improved efficiency. 

It is worth noting the distinction between our proposed sequence-level compression (inter-event) and standard representation-level reduction techniques (intra-event) commonly employed in efficient MLLMs, such as token pruning or merging \citep{bolya2022tome,chen2024image,dhouib2025pact,alvar2025divprune}. While these methods effectively reduce computational cost by exploiting spatial redundancy within individual modalities (e.g., visual patches), they are less applicable to the structured components of TPPs. 
Unlike images, key components of TPPs such as timestamps ($t_i$), event types ($e_i$), and text descriptions ($m_i$) contain dense semantic information. Every time value and word is precise and necessary for understanding the event sequence, so removing them risks losing critical logic. Therefore, we prioritize a sequence-level strategy that compresses temporally similar events. This approach effectively captures the bursty nature of TPPs and allows the model to accommodate a much longer history within the limited context window. We leave the design of more advanced, hybrid compression strategies that adapt to more complex and diverse scenarios for future work.

\subsection{Training and Inference}

\paragraph{Model Training.}  
To enable the model to understand complex multimodal event structures and the proposed compression strategy, we adopt a two-stage training framework inspired by Language-TPP. In the first stage, continued pre-training is conducted on large-scale token sequences constructed from the defined multimodal event templates, which may include both full and compressed representations. This stage helps the model adapt to the new event format and learn the semantics of special tokens such as \texttt{<|similar\_event|>} and \texttt{<|type\_X|>}. The objective follows standard next-token prediction:
\begin{equation}
\mathcal{L}_{\text{stage1}}(\theta) = -\frac{1}{L} \sum_{i=1}^{L} \log p_{\theta}(x_i \mid x_{<i}),
\end{equation}
where $x_1, \ldots, x_L$ denotes the event token sequence.

The second stage involves supervised fine-tuning for downstream tasks, including time prediction, type classification, and text generation. Training samples are formatted into prompt-response pairs, where the prompt consists of a compressed historical segment and a task-specific instruction token (e.g., \texttt{<|time\_prediction|>}, \texttt{<|type\_prediction|>}, or a natural language query), and the response contains the ground-truth tokens (e.g., byte tokens for time, type token, or text sequence). The training loss is applied only to the response tokens:
\begin{equation}
\mathcal{L}_{\text{stage2}}(\theta) = -\frac{1}{R} \sum_{j=1}^{R} \log p_{\theta}(r_j \mid \text{Prompt}, r_{<j}),
\end{equation}
where $r_1, \ldots, r_R$ denote the response sequence. This two-stage process enables the model to learn both general sequence modeling and task-specific reasoning over multimodal event data. 

\paragraph{Model Inference.}  
During inference, the model follows an autoregressive decoding strategy similar to Language-TPP. Given a historical sequence and a task-specific instruction, it sequentially generates output tokens. Byte tokens are decoded into floating-point intervals for time prediction, categorical tokens indicate event types, and language tokens are interpreted as text for generation or QA tasks. Since Qwen2.5-VL~\citep{bai2025qwen2} does not support image generation, our current model focuses on time, type, and text outputs. Future extensions may leverage omni-modal models such as Chameleon~\citep{chameleonteam2025chameleonmixedmodalearlyfusionfoundation} to enable image generation for predicted events.

\section{Experiments}\label{setup}
We conduct comprehensive experiments to evaluate the effectiveness of our proposed MM-TPP across multiple dimensions, including overall performance against SOTA baselines, text generation quality, the impact of adaptive compression mechanism, and the contribution of multimodal information integration.

\subsection{Experimental Setup}

Our MM-TPP model is built upon the Qwen2.5-VL-3B architecture and fine-tuned using LoRA (Low-Rank Adaptation)~\citep{hu2021loralowrankadaptationlarge}. All experiments are conducted on a single NVIDIA RTX 4090 GPU. For Stage 1 (continued pre-training), we train the model for 5 epochs with a learning rate of $1 \times 10^{-4}$. In Stage 2 (supervised fine-tuning), we also use a learning rate of $1 \times 10^{-4}$ for 3 epochs. The temporal similarity threshold for our compression mechanism is set to $\Delta = 0.2$, which we found to offer a good balance between compression and performance. During inference, we set the decoding temperature to 0.05 and perform each evaluation 5 times with different random seeds to ensure robustness. Full hyperparameter details can be found in \cref{ap:hp}.

\paragraph{Baselines.}  
We compare MM-TPP with several SOTA TPP models, including \textbf{NHP}~\citep{mei2017neural}, \textbf{SAHP}~\citep{zhang2020self}, \textbf{THP}~\citep{simiao2020transformer}, \textbf{RMTPP}~\citep{du2016recurrent}, \textbf{AttNHP}~\citep{mei2022transformer}, \textbf{TPP-LLM}~\citep{liu2024tpp}, \textbf{S2P2}~\citep{chang2025deep},and \textbf{Language-TPP}~\citep{kong2025language}. All models except TPP-LLM and Language-TPP are trained and evaluated using the Easy-TPP framework~\citep{xue2024easytpp}, while TPP-LLM and Language-TPP are reproduced using its official implementation. For a fair comparison, we use a unified maximum epoch setting across all baselines, select the best-performing checkpoints for testing, and repeat the entire training and evaluation pipeline 5 times with different random seeds.

\paragraph{Datasets.}
Most existing TPP benchmarks are limited to timestamps and event types, with very few incorporating both textual and visual data. We evaluate MM-TPP on two distinct multimodal datasets to address this gap:

\begin{itemize}[leftmargin=*]
    \item \textbf{DanmakuTPP}~\citep{jiang2025danmakutppbench}. An established dataset that contains extremely long sequences—most with over 1,000 events, and some exceeding 9,600. Each event includes a timestamp, textual content (e.g., video comments), and visual information extracted from corresponding video frames. It also provides the \textbf{DanmakuTPP-QA} subset for assessing complex multimodal reasoning capabilities.

    \item \textbf{TAXI-PRO}. A new dataset we introduce by augmenting the classic NYC Taxi dataset~\citep{whong2014foiling} with rich multimodal content. To create a complementary evaluation scenario, we enrich each event with $224 \times 224$ map image patches centered at each GPS coordinate, as well as natural language descriptions incorporating landmarks, coordinates, and ride metadata (see \cref{app:taxi} for details). In contrast to DanmakuTPP, sequences in TAXI-PRO are significantly shorter (average length of 36), providing a complementary evaluation scenario where our proposed compression mechanism is not required.
\end{itemize}

\paragraph{Metrics.}  
We evaluate model performance using several task-specific metrics. For timestamp prediction, we report root mean square error (\textbf{RMSE}); for event type classification, we use accuracy (\textbf{ACC}); and for text generation, we measure token-level quality using perplexity (\textbf{PPL}). For question answering tasks in DanmakuTPP-QA, we follow the original evaluation protocol—using \textbf{ACC} and \textbf{RMSE} for closed-ended questions, and conducting \textbf{qualitative analysis} for open-ended responses.

\begin{table}[t]
\centering
\caption{Prediction performance comparison on multimodal TPP datasets for event times and types. Results reported in terms of RMSE($\downarrow$) / ACC($\uparrow$) with standard deviations. Best results are in \textbf{bold}.}
\label{tab:results}
\resizebox{0.8\textwidth}{!}{
\begin{tabular}{lcccc}
\toprule
\multirow{2}{*}{\textbf{Model}} & \multicolumn{2}{c}{\textbf{DanmakuTPP}} & \multicolumn{2}{c}{\textbf{TAXI-PRO}} \\
\cmidrule(lr){2-3} \cmidrule(lr){4-5}
& \textbf{RMSE}($\downarrow$) & \textbf{ACC(\%)}($\uparrow$) & \textbf{RMSE}($\downarrow$) & \textbf{ACC(\%)}($\uparrow$) \\
\midrule
NHP & 5.4540 {\scriptsize $\pm$0.036} & 30.74 {\scriptsize $\pm$0.08} & 0.4494 {\scriptsize $\pm$0.012} & 75.93 {\scriptsize $\pm$0.20} \\
SAHP & 5.4245 {\scriptsize $\pm$0.006} & 27.38 {\scriptsize $\pm$0.05} & 0.3526 {\scriptsize $\pm$0.006} & 75.37 {\scriptsize $\pm$1.30} \\
THP & 5.4001 {\scriptsize $\pm$0.002} & 24.64 {\scriptsize $\pm$0.06} & 0.3736 {\scriptsize $\pm$0.008} & 75.31 {\scriptsize $\pm$1.00} \\
RMTPP & 5.4633 {\scriptsize $\pm$0.076} & 23.82 {\scriptsize $\pm$0.50} & 0.3830 {\scriptsize $\pm$0.008} & 76.47 {\scriptsize $\pm$0.20} \\
AttNHP & 5.4384 {\scriptsize $\pm$0.077} & 28.92 {\scriptsize $\pm$0.40} & 0.4049 {\scriptsize $\pm$0.007} & 69.37 {\scriptsize $\pm$0.80} \\
S2P2 & 5.5490 {\scriptsize $\pm$0.016} & \textbf{31.48} {\scriptsize $\pm$0.13} & 0.5735 {\scriptsize $\pm$0.035} & 75.73 {\scriptsize $\pm$0.60} \\
\midrule
TPP-LLM & 5.3035 {\scriptsize $\pm$0.012} & 24.59 {\scriptsize $\pm$0.70} & 0.3336 {\scriptsize $\pm$0.008} & 71.09 {\scriptsize $\pm$0.20} \\
Language-TPP & 5.3845 {\scriptsize $\pm$0.022} & 22.62 {\scriptsize $\pm$0.07} & 0.3376 {\scriptsize $\pm$0.001} & 75.27 {\scriptsize $\pm$0.08} \\
MM-TPP & \textbf{5.2987} {\scriptsize $\pm$0.018} & 27.62 {\scriptsize $\pm$0.02} & \textbf{0.3310} {\scriptsize $\pm$0.001} & \textbf{77.56} {\scriptsize $\pm$0.04} \\
\bottomrule
\end{tabular}}
\end{table}

\subsection{Overall Performance Comparison}
We first compare MM-TPP against all baseline methods on the complete datasets. Both datasets are split into training, validation, and test sets with an 8:1:1 ratio. 
For DanmakuTPP, we constrain the number of encoded events using a maximum token limit of 4096 tokens during training and perform evaluation on the first 200 events of each test sequence. 
For TAXI-PRO, the 4096 token limit is sufficient to encode complete sequences without truncation. 
As shown in \cref{tab:results}, MM-TPP delivers state-of-the-art performance across both datasets. On DanmakuTPP, it achieves the lowest RMSE (\textbf{5.2987}) for time prediction, while its type accuracy (27.62\%) is highly competitive and substantially improves upon Language-TPP. The advantages are even clearer on TAXI-PRO, where MM-TPP outperforms all baselines on both metrics, establishing a new benchmark with a leading RMSE of \textbf{0.3310} and an accuracy of \textbf{77.56\%}.

\subsection{Performance on Multimodal Question Answering}
A key strength of our framework lies in its capacity for complex reasoning over multimodal event streams. To evaluate this, we adopt the \textbf{DanmakuTPP-QA} benchmark~\citep{jiang2025danmakutppbench}, which is specifically designed to assess temporal, visual, and textual reasoning. The benchmark comprises 8 closed-ended tasks for quantitative evaluation and 2 open-ended tasks that require generating descriptive analytical reports, evaluated qualitatively. Detailed descriptions of all 10 tasks are provided in \cref{ap:QA}.


\begin{table}[t]
\centering
\small 
\caption{Comparative evaluation on the DanmakuTPP-QA closed-ended tasks. Metrics are ACC ($\uparrow$) and RMSE ($\downarrow$). The proposed MM-TPP model is highlighted. Best results are in \textbf{bold} and the second best results are set in \underline{underline}.}
\resizebox{\linewidth}{!}{
\begin{tabular}{lcccccccc}
\toprule
\textbf{Model / Task} & \textbf{T-1} & \textbf{T-2} & \textbf{T-3} & \textbf{T-4} & \textbf{T-5} & \textbf{T-6} & \textbf{T-7} & \textbf{T-8} \\
\textbf{Metrics} & ACC~$\uparrow$ & RMSE~$\downarrow$ & RMSE~$\downarrow$ & RMSE~$\downarrow$ & RMSE~$\downarrow$ & RMSE~$\downarrow$ & ACC~$\uparrow$ & ACC~$\uparrow$ \\
\midrule
\multicolumn{9}{l}{\textit{\textbf{LLMs}}} \\
Qwen2.5-7B-Instruct & 0.33 & 27.64 & 134.45 & 0.65 & 0.56 & 0.51 & 10.67 & 32.67 \\
Qwen2.5-32B-Instruct & 25.33 & 1.52 & 122.69 & 0.36 & 0.29 & 0.24 & 16.67 & 38.17 \\
Qwen2.5-72B-Instruct & 0.67 & 1.28 & 123.45 & 0.30 & 0.46 & 0.46 & 16.00 & 43.83 \\
Qwen3-8B & 6.67 & 1.80 & 123.59 & 0.32 & 0.41 & 0.45 & \underline{19.33} & 41.50 \\
Qwen3-30B-A3B & 0.67 & 1.33 & 121.96 & \underline{0.20} & 0.33 & 0.40 & \textbf{23.00} & 43.67 \\
Qwen3-235B-A22B & 8.67 & 1.39 & \textbf{120.79} & 0.30 & 0.31 & 0.29 & 10.33 & 32.50 \\
Llama-3.3-70B-Instruct & 1.67 & \textbf{1.11} & 121.49 & 0.26 & 0.27 & 0.22 & 17.00 & 33.33 \\
DeepSeek-V3 & 25.00 & 1.30 & 121.30 & 0.34 & 0.26 & 0.22 & 13.67 & 34.5 \\
\midrule
\multicolumn{9}{l}{\textit{\textbf{MLLMs}}} \\
Qwen2.5-VL-7B & 9.67 & 11.61 & 124.99 & 0.46 & 0.82 & 0.66 & 8.33 & 22.17 \\
Qwen2.5-VL-32B & 8.0 & 1.26 & 124.02 & 0.35 & 0.51 & 0.38 & 12.67 & 22.17 \\
Qwen2.5-VL-72B & 0.33 & \underline{1.14} & \underline{121.25} & 0.28 & 0.47 & 0.41 & 15.98 & \textbf{47.17} \\
Gemma3-27B & 0.33 & 1.33 & 121.32 & 0.28 & 0.27 & 0.20 & 15.67 & 36.17 \\
\midrule
\multicolumn{9}{l}{\textit{\textbf{Finetuned}}} \\
Qwen2.5-VL-3B & \underline{27.0} & 1.35 & 220.43 & \textbf{0.05} & \underline{0.16} & \underline{0.08} & 15.33 & 43.00 \\
\hline 
\rowcolor{yellow!50}
\textbf{MM-TPP} & \textbf{27.33}& 1.49& 190.45&\textbf{0.05} & \textbf{0.10}& \textbf{0.07}&15.67 &\underline{44.00} \\
\bottomrule
\end{tabular}}
\label{tab-exp-closed}
\end{table}

\paragraph{Closed-ended QA Tasks.} We compare MM-TPP with SOTA LLMs and MLLMs on eight closed-ended tasks, using baseline results reported in DanmakuTPPBench~\citep{jiang2025danmakutppbench}. As shown in \cref{tab-exp-closed}, MM-TPP achieves highly competitive performance across the benchmark and attains leading results on several tasks. Notably, it consistently outperforms the fine-tuned Qwen2.5-VL-3B from the original benchmark, despite utilizing the same underlying base model. This performance gap highlights the strength of our framework, which explicitly captures the structured nature of multimodal event sequences—unlike conventional fine-tuning approaches. 



\paragraph{Open-ended QA and Report Generation.}
We assess MM-TPP’s capacity for generating high-quality analytical text—a core capability that sets it apart from purely predictive models. On the two open-ended tasks from the DanmakuTPP-QA benchmark, MM-TPP consistently produces significantly more insightful and coherent text compared to existing MLLMs. To further evaluate its generalization ability, we prompt MM-TPP to generate descriptive analytical reports on the TAXI-PRO dataset, which lacks predefined QA tasks and has not been used to fine-tune the model for generative tasks. Remarkably, the quality of the generated narratives for TAXI-PRO’s temporal data also surpasses that of leading MLLMs, which are often limited to basic descriptive analysis of the event sequence. In contrast, MM-TPP uncovers high-level spatio-temporal patterns, demonstrating a robust capacity to synthesize multimodal event streams into coherent, insightful narratives—even in a zero-shot setting. Detailed qualitative comparisons of MM-TPP and baseline MLLM outputs for both datasets are provided in \cref{ap:QA}.

\subsection{Effectiveness of Compression Strategy}
To assess the effectiveness of our adaptive compression strategy, we compare the performance of MM-TPP with and without this mechanism on the DanmakuTPP dataset (note that compression is not applied to TAXI-PRO due to its shorter sequences). The key advantage of compression lies in its ability to fit longer event histories within a fixed context window. Under a 4096-token constraint, the uncompressed model can accommodate an average of 113 events, whereas the compressed variant extends this to an average of 292 events, with a maximum of 2008. This extended context allows the model to better capture long-range dependencies, leading to noticeable performance gains. As a result, the compressed model outperforms its uncompressed counterpart, achieving lower RMSE (5.2987 vs. 5.5551) and higher accuracy (27.62\% vs. 25.87\%) on predictive tasks. To further verify the effectiveness of compression strategy, we compared MM-TPP against a naive baseline that randomly drops events. As detailed in Appendix~\ref{app:random_drop}, the random drop strategy leads to significant performance degradation, confirming that preserving temporal causality is as critical as extending context.

\begin{figure}[t]
    \centering 
    \begin{subfigure}[b]{0.32\textwidth}
        \centering
        \includegraphics[width=\linewidth]{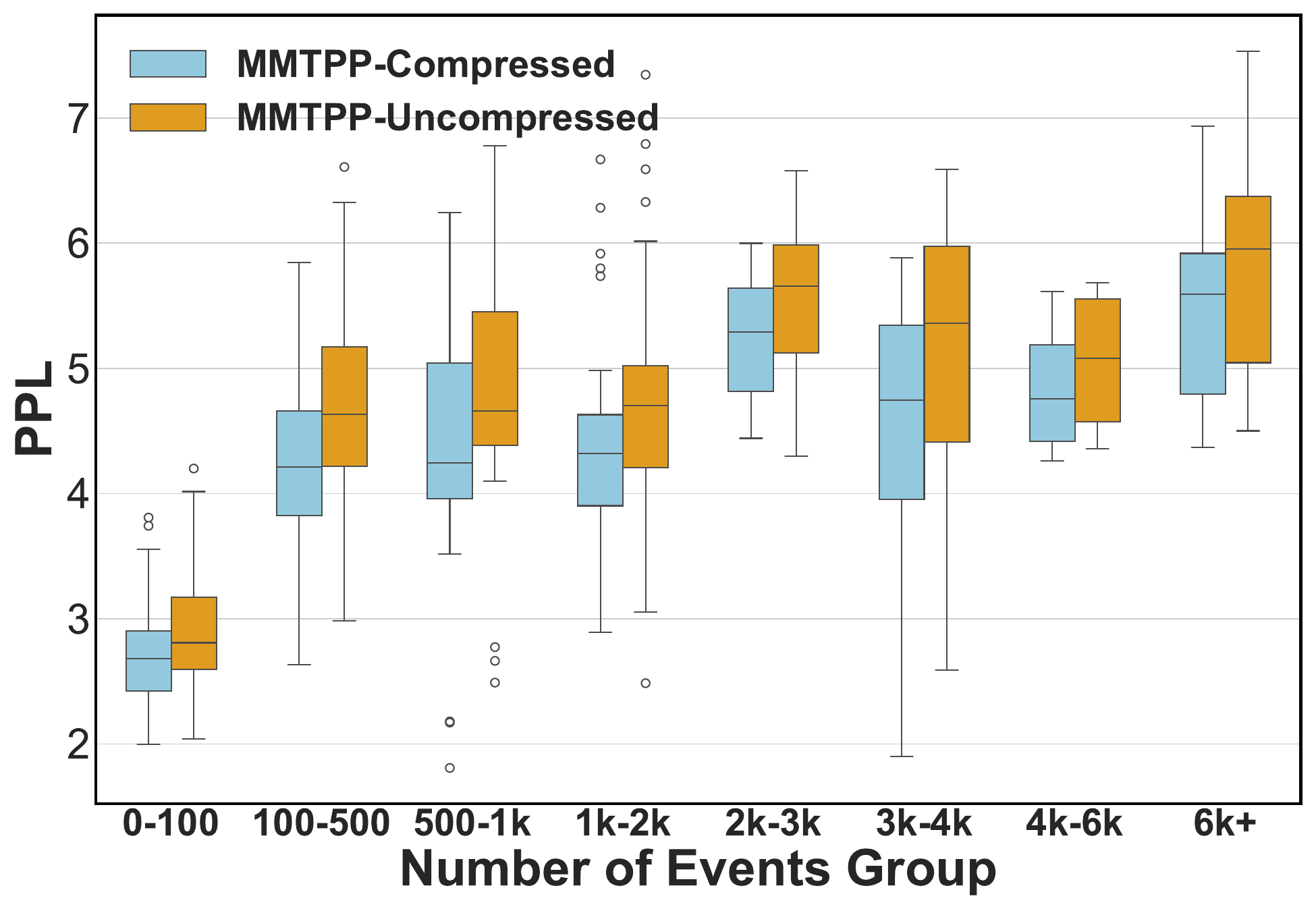} 
        \caption{}
        \label{fig:ppl}
    \end{subfigure}
    \hfill 
    \raisebox{-0.2cm}{
    \begin{subfigure}[b]{0.32\textwidth}
        \centering
        \includegraphics[width=\linewidth]{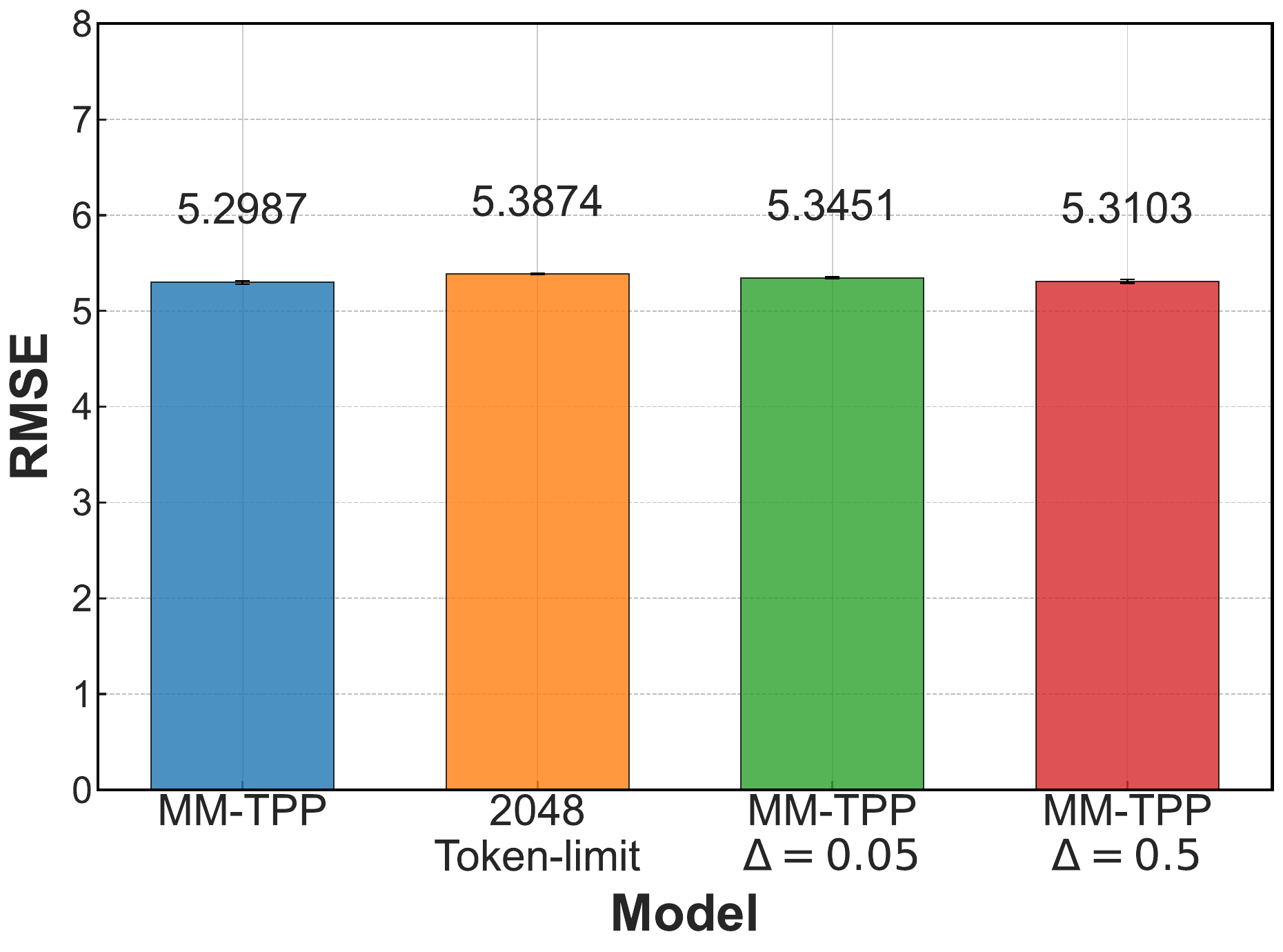}
        \caption{}
        \label{fig:placeholder1_side}
    \end{subfigure}%
}
    \hfill 
    \raisebox{-0.2cm}{
    \begin{subfigure}[b]{0.32\textwidth}
        \centering
        \includegraphics[width=\linewidth]{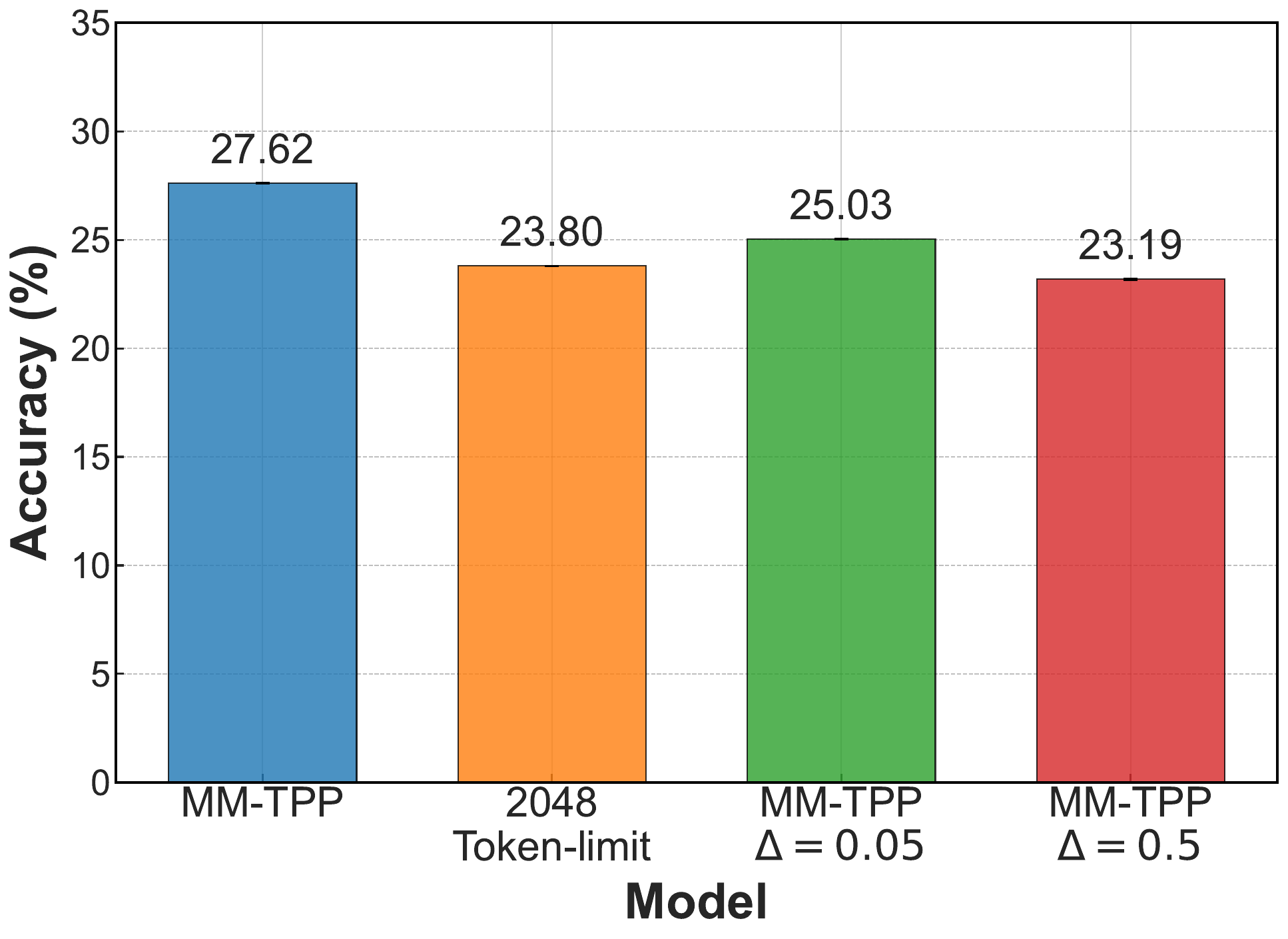} 
        \caption{}
        \label{fig:placeholder2_side}
    \end{subfigure}
}
    
    \caption{Evaluation of the compression mechanism and key hyperparameters on the DanmakuTPP dataset. (a) PPL comparison on the test set. Our compressed model consistently maintains a lower PPL than the uncompressed version, indicating effective long-range modeling. (b, c) Ablation results for context length and similarity threshold ($\Delta$) on RMSE and ACC, showing optimal performance with a larger context and $\Delta=0.2$.}
    \label{fig:full_row_grid}
\end{figure}

In addition, we conduct further experiments to evaluate the effectiveness of our compression method from a language modeling perspective. Specifically, we measure \textbf{PPL} on the encoded event sequences from the test set without truncation, a standard and widely-used metric for assessing the quality of language models \citep{zhao2023survey}. As shown in \cref{fig:ppl}, perplexity increases with sequence length—a well-known challenge in sequence modeling. Notably, our compressed model consistently achieves lower PPL across all lengths, with the gap widening as sequences grow longer. These results highlight a key strength of our approach: by preserving more long-range context within a fixed token budget, the compression mechanism mitigates performance degradation on long and complex event streams.

\subsection{Ablation Study}

In this section, we present ablation studies to examine the impact of four key design choices in MM-TPP: (i) the incorporation of visual information, (ii) model scale, (iii) context window size (controlled via the maximum token limit), and (iv) the temporal similarity threshold used in the compression mechanism. 

\paragraph{Impact of Visual Information.}
To assess the contribution of visual input, we compare the full MM-TPP model (text + image) with two text-only variants: MM-TPP (text only) and Language-TPP. All models are evaluated under controlled conditions with a fixed number of events in the context window, ensuring that sequence length does not confound the results. As shown in \cref{tab:ablation}, MM-TPP consistently outperforms both baselines across datasets, indicating that visual features provide additional predictive value. The integration of image improves both timestamp and type prediction accuracy, highlighting the complementary role of visual context alongside textual information. 

\begin{table}[t]
\centering
\caption{Ablation study results for MM-TPP. The table evaluates the contribution of visual information (vs. text-only variants) and the effect of model size (vs. a 7B model). Metrics are RMSE($\downarrow$) and ACC($\uparrow$). Best results are in \textbf{bold}.}
\label{tab:ablation}
\resizebox{.8\textwidth}{!}{
\begin{tabular}{lcccc}
\toprule
\multirow{2}{*}{\textbf{Model}} & \multicolumn{2}{c}{\textbf{DanmakuTPP}} & \multicolumn{2}{c}{\textbf{TAXI-PRO}} \\
\cmidrule(lr){2-3} \cmidrule(lr){4-5}
& \textbf{RMSE}($\downarrow$) & \textbf{ACC(\%)}($\uparrow$) & \textbf{RMSE}($\downarrow$) & \textbf{ACC(\%)}($\uparrow$) \\
\midrule
MM-TPP & 5.2987 {\scriptsize $\pm$0.018} & \textbf{27.62} {\scriptsize $\pm$0.02} & \textbf{0.3310} {\scriptsize $\pm$0.001} & \textbf{77.56} {\scriptsize $\pm$0.04} \\
Language-TPP & 5.3845 {\scriptsize $\pm$0.022} & 22.62 {\scriptsize $\pm$0.07} & 0.3376 {\scriptsize $\pm$0.001} & 75.27 {\scriptsize $\pm$0.08} \\
MM-TPP (text only) & 5.4654 {\scriptsize $\pm$0.008} & 23.64 {\scriptsize $\pm$0.06} & 0.3388 {\scriptsize $\pm$0.001} & 76.70 {\scriptsize $\pm$0.08} \\
MM-TPP (7B) & \textbf{5.0533} {\scriptsize $\pm$0.002} & 26.98 {\scriptsize $\pm$0.01} & 0.3337 {\scriptsize $\pm$0.001} & 76.16 {\scriptsize $\pm$0.05}\\
\bottomrule
\end{tabular}}
\end{table}


\paragraph{Impact of Model Size.}
To assess the role of model capacity, we replace the base model Qwen2.5-VL-3B with a larger Qwen2.5-VL-7B, and evaluate on DanmaluTPP and TAXI-PRO datasets using identical training procedures and hyperparameters. This experiment investigates whether scaling up the model improves event prediction performance. Results in \cref{tab:ablation} show that scaling to 7B yields mixed results. While the larger model significantly improves time prediction (RMSE) on the complex DanmakuTPP dataset, the original 3B model outperforms it across all other metrics. This suggests a larger model's capacity is most beneficial for long-range temporal patterns but may hinder performance on simpler tasks, possibly due to overfitting.

\paragraph{Impact of Context Length.}
To examine the effect of context window size, we reduce the maximum token limit from 4096 to 2048 during both training and inference on the DanmakuTPP dataset. As shown in \cref{fig:placeholder1_side,fig:placeholder2_side}, this reduction results in a clear drop in performance. A shorter context window limits the model’s access to historical events, hindering its ability to capture long-range dependencies critical for accurate temporal modeling. These results highlight the importance of maintaining a sufficiently large context window when dealing with long and complex event sequences.

\paragraph{Impact of Temporal Similarity Threshold.}
\label{similartoken}
We evaluate how the performance of our compression mechanism varies with the temporal similarity threshold $\Delta$. Specifically, we compare the default setting $\Delta = 0.2$ with a stricter threshold ($\Delta = 0.05$) and a looser one ($\Delta = 0.5$), as motivated in \cref{ap:similar}. As shown in \cref{fig:placeholder1_side,fig:placeholder2_side}, both alternatives lead to performance drops. A small threshold ($\Delta = 0.05$) results in limited compression, offering little gain in contextual coverage. In contrast, a large threshold ($\Delta = 0.5$) causes over-compression, merging dissimilar events and discarding important temporal signals. These results confirm that $\Delta = 0.2$ strikes a good balance between compression efficiency and temporal fidelity.

\section{Conclusion}
In this paper, we introduce MM-TPP, a novel framework that advances TPP modeling from a purely predictive to a generative paradigm. By incorporating visual information alongside temporal, categorical, and textual data, our model can not only predict future events but also generate rich, descriptive text that analyzes and explains event dynamics. We address the critical challenge of long-context modeling in multimodal sequences through an adaptive compression strategy that exploits temporal similarity. Extensive experiments demonstrate that MM-TPP consistently outperforms SOTA baselines in standard prediction tasks and, more importantly, exhibits superior text generation and reasoning capabilities on the challenging DanmakuTPP-QA benchmark. While our framework currently lacks image generation, it opens new avenues for vision-enhanced temporal modeling and provides a solid foundation for future research in generative multimodal TPPs.

\section*{Acknowledgments}
This work was supported by the NSFC Project (No.62576346), the MOE Project of Key Research Institute of Humanities and Social Sciences (22JJD110001), the fundamental research funds for the central universities, and the research funds of Renmin University of China (24XNKJ13), and Beijing Advanced Innovation Center for Future Blockchain and Privacy Computing.

\section*{Ethics Statement}
Our work adheres to the ICLR Code of Ethics. The datasets used are either public benchmarks (DanmakuTPP) or derived from public data (NYC TLC). For our constructed TAXI-PRO dataset, we ensured no personally identifiable information was included, focusing only on aggregated spatio-temporal patterns. We acknowledge the potential for biases inherited from the pre-trained language model backbone. Our framework is intended for research purposes to analyze event dynamics, and we foresee no direct negative societal consequences from its use.

\section*{Reproducibility Statement}
We have made every effort to ensure the reproducibility of our research. The complete source code for our MM-TPP model, including training and evaluation procedures, will be provided in the supplementary materials. All crucial hyperparameters for both the pre-training and supervised fine-tuning stages are documented in \cref{ap:hp}. Furthermore, details of our experimental setup, baselines, and evaluation metrics are described in \cref{setup} to facilitate fair comparison and replication of our results.

\bibliography{iclr2026_conference}
\bibliographystyle{iclr2026_conference}

\newpage
\appendix
\section{Details of Adaptive Sequence Compression}
\subsection{Illustration of Compression Mechanism}
\label{app:compression_mechanism}
To clarify how the adaptive compression operates on consecutive events, we provide a step-by-step illustration using a hypothetical sequence of three events: $A$, $B$, and $C$, occurring in that order. Let $\tau_i = t_i - t_{i-1}$ denote the inter-event interval for event $i$. The compression decision is made sequentially based on the pairwise comparison of adjacent intervals:

\begin{itemize}
    \item \textbf{Step 1 (Processing Event B):} The model compares the interval of event $B$ ($\tau_B$) with the interval of the immediately preceding event $A$ ($\tau_A$). If the absolute difference satisfies $|\tau_B - \tau_A| < \Delta$, event $B$ is considered temporally similar to $A$ and is replaced by a single special token \texttt{<|similar\_event|>}.
    
    \item \textbf{Step 2 (Processing Event C):} The model then proceeds to event $C$. It compares its interval ($\tau_C$) with the interval of the preceding event $B$ ($\tau_B$). Note that this comparison is based on the calculated time intervals regardless of whether $B$ was compressed in the token sequence. If $|\tau_C - \tau_B| < \Delta$, event $C$ is also replaced by the \texttt{<|similar\_event|>} token.
\end{itemize}

This sequential process ensures that compression is applied locally based on the dynamic evolution of event density, allowing the model to efficiently encode bursts of events with similar distinct timing patterns.

\subsection{Temporal Similarity Threshold Setting}\label{ap:similar}

In our framework, the temporal similarity threshold $\Delta$ is a key hyperparameter for adaptive sequence compression. Specifically, for each event, we define the interval $\tau_i = t_i - t_{i-1}$ and compare it with the previous interval $\tau_{i-1}$. If the absolute difference between the two intervals satisfies $|\tau_i - \tau_{i-1}| < \Delta$, event $i$ is deemed temporally similar to event $i-1$ and is eligible for merging. This mechanism directly controls the granularity of sequence modeling and the preservation of fine-grained temporal dynamics, as only sufficiently similar events are compressed into a single representation.

\begin{figure}[H]
    \centering
    \begin{minipage}[b]{0.48\textwidth}
        \centering
        \includegraphics[width=\textwidth]{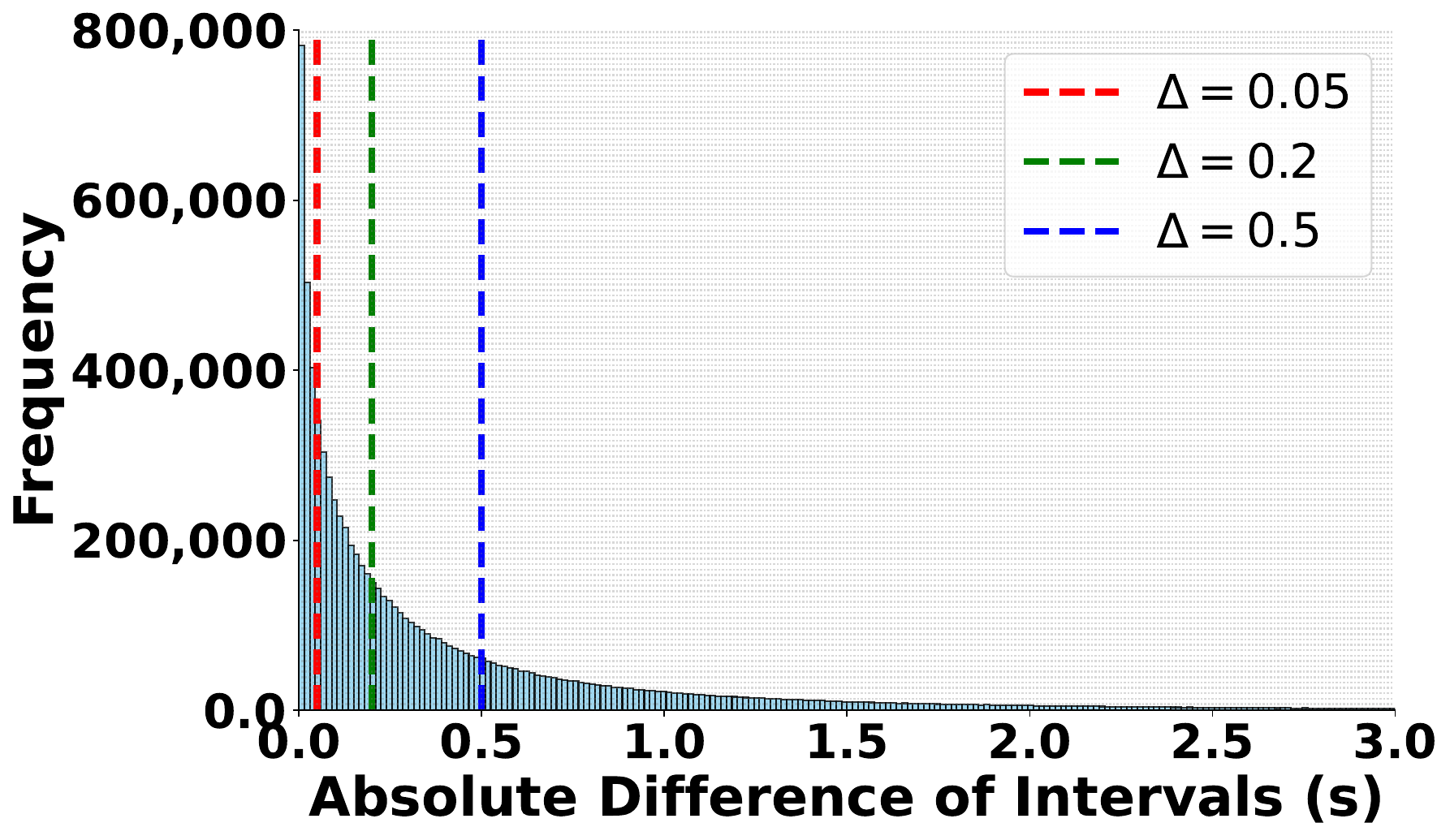}
        \caption{Distribution of the absolute difference between adjacent inter-event intervals on the training set. Vertical lines indicate the tested thresholds $\Delta$.}
        \label{fig:interval_dist}
    \end{minipage}
    \hfill 
\begin{minipage}[b]{0.48\textwidth}
\centering
\begin{table}[H]
\caption{Quantiles of the absolute difference between adjacent inter-event intervals ($|\tau_i - \tau_{i-1}|$).}
\label{tab:quantiles}
\centering
\begin{tabular}{cc}
\toprule
Percentile & Value (s) \\
\midrule
0.050 & 0.007 \\
0.100 & 0.017 \\
0.200 & 0.045 \\
0.250 & 0.063 \\
0.500 & 0.214 \\
0.750 & 0.590 \\
0.900 & 1.330 \\
0.950 & 2.115 \\
1.000 & 1787.000 \\
\bottomrule
\end{tabular}
\end{table}
    \end{minipage}
\end{figure}

To determine an appropriate value for $\Delta$, we conducted a comprehensive statistical analysis of the absolute differences between adjacent inter-event intervals ($|\tau_i - \tau_{i-1}|$) on the DanmakuTPP training dataset. As illustrated in \cref{fig:interval_dist}, the empirical distribution of these differences is highly skewed, with a large proportion of values concentrated near zero. This indicates that event intervals are frequently very similar to their predecessors, validating the potential for a compression strategy based on temporal similarity.

To ground our threshold selection in the data's structure, we examined the distribution's quantiles, summarized in \cref{tab:quantiles}. Based on this analysis, we selected candidate thresholds of 0.05, 0.2, and 0.5 seconds. These values correspond to meaningful points in the distribution:
\begin{itemize}
    \item $\Delta = 0.05$ is a conservative threshold, corresponding to approximately the 25th percentile. It compresses only the most similar event pairs, preserving most of the original sequence structure.
    \item $\Delta = 0.2$ corresponds to approximately the 50th percentile (the median), offering a balance between compression rate and information preservation.
    \item $\Delta = 0.5$ is a more aggressive threshold, near the 75th percentile, which merges a larger portion of the sequence to achieve a higher compression ratio.
\end{itemize}
The positions of these thresholds relative to the distribution are marked in \cref{fig:interval_dist}. This principled, data-driven selection ensures that our adaptive compression mechanism is both statistically grounded and practically effective for multimodal temporal point process modeling.

\section{TAXI-PRO Dataset Construction}\label{app:taxi}

We construct the TAXI-PRO dataset by augmenting the original NYC Taxi dataset with multimodal information (images and text descriptions). This section provides detailed information about the data construction pipeline.

\paragraph{Original Taxi Data.} The raw taxi trip data is sourced from the New York City Taxi and Limousine Commission (NYC TLC) public dataset, consisting of files \texttt{trip\_data\_1.csv} through \texttt{trip\_data\_12.csv}. This dataset is identical to the source used by the EasyTPP official taxi dataset and originates from NYC TLC's open data initiative~\footnote{FOILing NYC's Taxi Trip Data: \url{https://chriswhong.com/open-data/foil_nyc_taxi/}}.

We construct the TAXI-PRO dataset with the following setups:
\begin{itemize}
    \item \textbf{Image}. We leverage the coordinates of each event and construct an image crop of the map of the location. High-resolution map tiles are generated from OpenStreetMap data using tools like \texttt{contextily}. The base map covers the entire Manhattan area with strict correspondence between pixel coordinates and geographical boundaries (resolution: 6071×6307 pixels). For each event location, we extract a 224×224 pixel patch centered at the event coordinates. Edge cases are handled by padding with gray background to maintain consistent patch dimensions.
    \item \textbf{Text}. We generate the textual mark for each event following a fixed template incorporating landmark information, coordinates, passenger count, and trip distance. For example, \textit{"Picked up at Times Square (40.7580, -73.9855), 2 passengers"}.
    \item \textbf{Event type}. We focus exclusively on Manhattan and subdivide it into three regions: lower Manhattan, midtown Manhattan, and upper Manhattan, resulting in 6 event types for combinations of regions and pickup/dropoff events.
\end{itemize}

\begin{figure}[h!]
    \centering
    \includegraphics[width=0.8\linewidth]{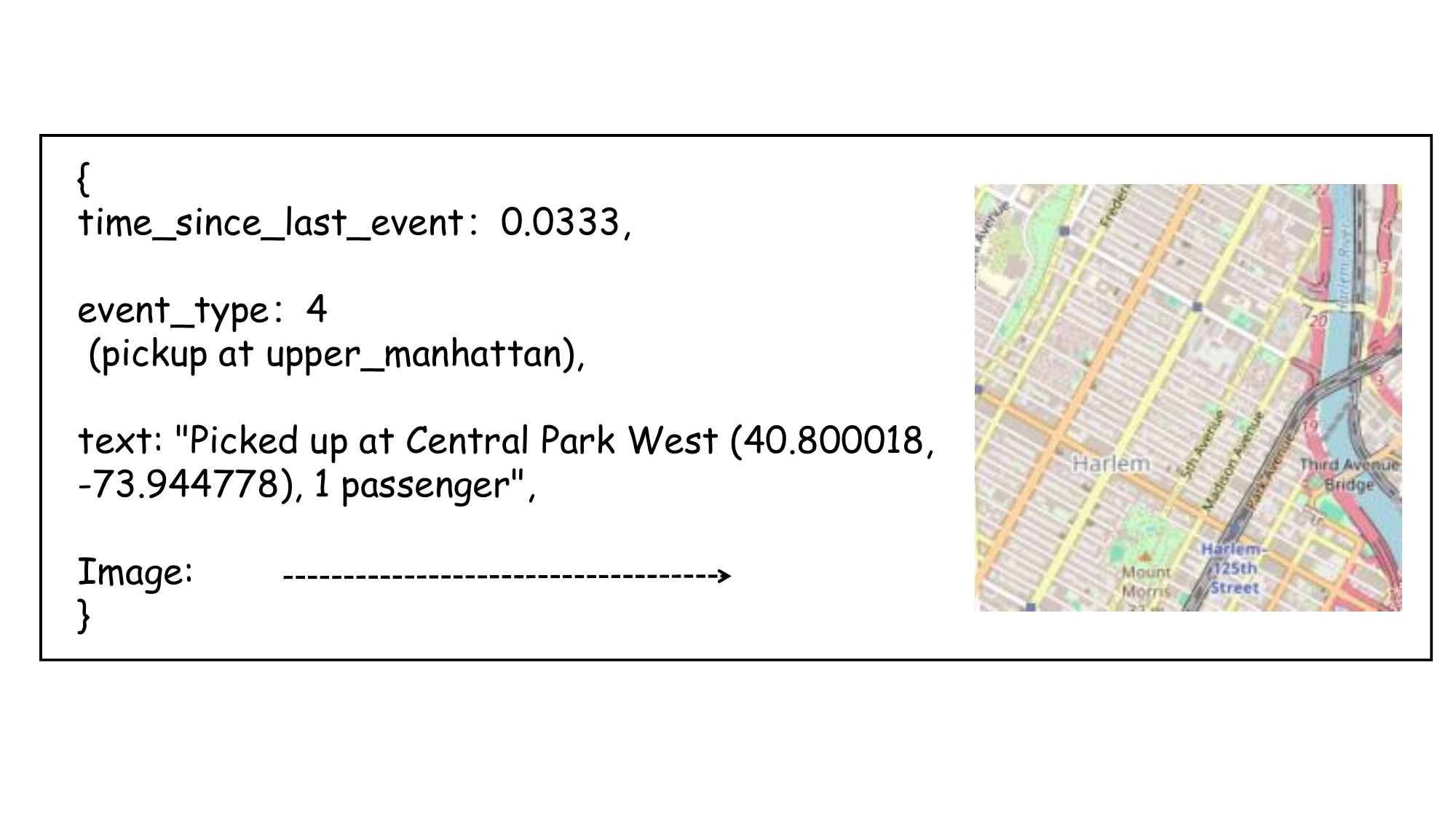}
    \caption{An illustration of a single event in the TAXI-PRO dataset. Each event comprises four key components: (1) a 224$\times$224 map image centered on the event's coordinates, (2) a generated textual description including landmark and trip details, (3) a categorical event type and (4) a timestamp.}
    \label{figure: taxi}
\end{figure}

To provide a concrete visualization of our data structure, \cref{figure: taxi} showcases the composition of a single event in the TAXI-PRO dataset. It highlights the multimodal nature of each data point, combining the geographical map patch, the templated text description, and the categorical event type.
When selecting the final 2000 sequences, we employ a greedy algorithm with variance minimization to ensure balanced event type distribution, avoiding imbalance.

\section{Detailed Hyperparameter Settings}\label{ap:hp}
\paragraph{Hyperparameters}
The specific hyperparameters used for each stage of our training procedure are summarized in \cref{table:hyper}. We employed the AdamW optimizer and utilized bfloat16 mixed-precision for all experiments to enhance computational efficiency.

\begin{table}[t]
\centering
\caption{Hyperparameter Setting}
\begin{tabular}{lc}
\toprule
Hyperparameter & Value \\
\midrule
Base Model & Qwen2.5-VL-3B \\
LoRA Rank & 16 \\
LoRA Alpha & 64 \\
LoRA Dropout & 0.1 \\
Stage 1 Learning Rate & 1e-4 \\
Stage 1 Epochs & 5 \\
Stage 2 Learning Rate & 1e-4 \\
Stage 2 Epochs & 3 \\
Batch Size & 1 \\
Max Token Length & 4096 \\
Temporal Threshold $\Delta$ & 0.2s \\
Inference Temperature & 0.05 \\
\bottomrule
\end{tabular}
\label{table:hyper}
\end{table}

\section{Comparison with Random Drop Baseline}
\label{app:random_drop}

To rigorously evaluate the effectiveness of our proposed adaptive compression mechanism, we compared it against a naive compression baseline: \textbf{Random Drop}. This baseline randomly discards events from the sequence during both training and inference phases.

We conducted experiments on the DanmakuTPP dataset with drop probabilities of $p=25\%$ and $p=50\%$. The results, compared with our MM-TPP (Adaptive Compression) and the uncompressed baseline, are presented in Table~\ref{tab:random_drop}.

\begin{table}[h]
    \centering
    \caption{Performance comparison between MM-TPP's adaptive compression and naive Random Drop strategies on the DanmakuTPP dataset. The proposed adaptive method significantly outperforms random dropping, highlighting the importance of preserving temporal structure.}
    \label{tab:random_drop}
    \vspace{2mm}
    \begin{tabular}{lcc}
    \toprule
    \textbf{Model / Setting} & \textbf{RMSE} ($\downarrow$) & \textbf{ACC (\%)} ($\uparrow$) \\
    \midrule
    MM-TPP & \textbf{5.2987}{\scriptsize $\pm$0.018} & \textbf{27.62}{\scriptsize $\pm$0.02} \\
    MMTPP-Uncompressed & 5.5551{\scriptsize $\pm$0.008} & 25.87{\scriptsize $\pm$0.05} \\
    \midrule
    MMTPP-Random-Drop ($p=25\%$) & 5.3792{\scriptsize $\pm$0.014} & 24.84{\scriptsize $\pm$0.06} \\
    MMTPP-Random-Drop ($p=50\%$) & 5.4259{\scriptsize $\pm$0.009} & 24.79{\scriptsize $\pm$0.03} \\
    \bottomrule
    \end{tabular}
\end{table}

As shown in Table~\ref{tab:random_drop}, the Random Drop strategy results in a clear performance degradation compared to our adaptive approach. 
Even with a moderate drop rate of 25\%, the model's predictive accuracy and timestamp precision are inferior to MM-TPP. This degradation is likely because event sequences in TPPs contain strict causal dependencies; randomly removing events disrupts the historical context necessary for the model to reason about future dynamics (e.g., self-exciting patterns). 
In contrast, our adaptive compression strategy selectively merges only temporally similar intervals (which often represent redundant burst signals) while preserving the critical anchor events and the overall semantic continuity of the sequence.

\section{Danmakutpp-QA Details}\label{ap:QA}
\begin{table}[h!]
\caption{Detailed descriptions and evaluation metrics for the ten tasks in the DanmakuTPP-QA benchmark. Closed-ended tasks are evaluated quantitatively (Accuracy or RMSE), while open-ended tasks are assessed via qualitative analysis.}
\centering
\resizebox{\textwidth}{!}{
\begin{tabular}{clcc}
\toprule
 & \textbf{Task Description} & \textbf{Evaluation Metrics} & \textbf{Task Type} \\
\midrule
Task-1 & \begin{tabular}[c]{@{}l@{}}Danmaku burst peak counting\end{tabular} & ACC & Closed-ended \\
\midrule
Task-2 & \begin{tabular}[c]{@{}l@{}}Prediction of the next\\ Danmaku timestamp\end{tabular} & RMSE & Closed-ended \\
\midrule
Task-3 & \begin{tabular}[c]{@{}l@{}}Prediction of the next Danmaku\\ burst peak timestamp\end{tabular} & RMSE & Closed-ended \\
\midrule
Task-4 & \begin{tabular}[c]{@{}l@{}}Assessment of average \\ sentiment polarity\end{tabular} & RMSE & Closed-ended \\
\midrule
Task-5 & \begin{tabular}[c]{@{}l@{}}Sentiment polarity prediction \\ for the next Danmaku\end{tabular} & RMSE & Closed-ended \\
\midrule
Task-6 & \begin{tabular}[c]{@{}l@{}}Sentiment polarity prediction\\ for the next Danmaku burst peak\end{tabular} & RMSE & Closed-ended \\
\midrule
Task-7 & \begin{tabular}[c]{@{}l@{}}Event type inference for\\ the next Danmaku\end{tabular} & ACC & Closed-ended \\
\midrule
Task-8 & \begin{tabular}[c]{@{}l@{}}Prediction of Top-2 triggering event\\ types for the next burst peak\end{tabular} & ACC & Closed-ended \\
\midrule
Task-9 & \begin{tabular}[c]{@{}l@{}}Analysis of global sentiment\\ dynamics and the underlying drivers\end{tabular} & Qualitative Analysis & Open-ended \\
\midrule
Task-10 & \begin{tabular}[c]{@{}l@{}}Causal attribution analysis for\\ specific Danmaku burst peak formation\end{tabular} & Qualitative Analysis & Open-ended \\
\bottomrule
\end{tabular}}
\label{tab:task_details}
\end{table}
\subsection{Ten Question-Answering Tasks}
This section provides a detailed breakdown of the ten reasoning tasks that constitute the \textbf{DanmakuTPP-QA} benchmark. The benchmark is designed to evaluate a model's temporal-visual-textual reasoning capabilities through a series of diverse challenges. As summarized in \cref{tab:task_details}, these tasks are divided into eight closed-ended, quantitatively evaluated tasks and two open-ended tasks that require qualitative assessment of generated reports.

\subsection{MM-TPP's performance on Open-ended Tasks}
In this section, we conduct a qualitative evaluation of MM-TPP's ability to generate high-quality analytical reports for open-ended tasks. We compare its outputs against two SOTA, general-purpose Multimodal Large Language Models (MLLMs): \textbf{Qwen2.5-VL-3B} \citep{bai2025qwen2} and \textbf{Gemma3-27B} \citep{team2025gemma}. Our analysis reveals that MM-TPP consistently produces significantly more insightful and relevant reports, highlighting the value of its specialized architecture.

The qualitative advantage is not only evident on the DanmakuTPP-QA benchmark, where MM-TPP was fine-tuned, but also generalizes robustly to the TAXI-PRO dataset in a zero-shot setting. To substantiate these claims, we provide a series of comparative examples in the following sections. Given the consistent nature of the performance gap, we present two representative examples for each of the open-ended tasks (Task 9 and Task 10) and for the TAXI-PRO report generation task to illustrate the consistent superiority of MM-TPP's analytical capabilities. Here we summarize the qualitative advantage as follows:
\begin{itemize}
    \item \textbf{Task 9 (Danmaku sentiment dynamics analysis)}: MM-TPP conveys information more effectively by focusing on essential sentiment dynamics, follows a clear analytical structure from overall trend to key inflection points and summary, and produces outputs that human evaluators find more readable and natural.
    \item \textbf{Task 10 (Danmaku peak burst reason analysis)}: Beyond capturing peak bursts, MM-TPP pinpoints sub-peaks with fine temporal precision, enabling a more rigorous view of self-exciting dynamics. It also ties specific event types, such as quote/reference and humor/meme, directly to the triggering mechanisms of bursts, clarifying how content categories drive cascading responses. Unlike the general-purpose MLLMs, it avoids drifting into tangential scene descriptions and instead highlights the key drivers, while framing the explanation explicitly in terms of TPP theory. This balance of precision, focus, and domain-specific grounding makes the analysis both accessible and academically robust.
    \item \textbf{TAXI-PRO}: MM-TPP can delivers high-level insights by identifying core TPP patterns in the driver's behavior. It moves beyond simple route tracking to model how events mutually influence one another, revealing strategic responses to spatio-temporal dynamics such as the formation of self-exciting event clusters in high-demand corridors while MLLMs, proficient in language generation, tend to produce outputs that are more descriptive than analytical.
\end{itemize}

\paragraph{TAXI-PRO Example\_1}
\begin{tcolorbox}[
  breakable,
  colback=lightgray!30,
  colframe=black!75,
  arc=0mm,
  boxrule=0.5pt,
  left=5pt,
  right=5pt,
  top=5pt,
  bottom=5pt,
  width=\textwidth,
  title={MM-TPP's generated text on TAXI-PRO (Example\_1)},
]
\scriptsize 
\textbf{Prompt:}\\
You are an expert data analyst specializing in spatio-temporal event sequences. Your task is to analyze a chronological log of a single taxi driver's activities in NYC, which includes actions like pickups and drop-offs at various locations. Based on the provided event sequence, please conduct a comprehensive analysis and generate a detailed Analysis in several paragraphs. Note that this sequence is NOT VIDEO! All time intervals are converted to four bytes according to the IEEE 754 standard.

$<|TPP\_data|>$

\textbf{Answer:}\\
The event sequence reveals a driver’s dynamic movement across multiple iconic New York City (NYC) neighborhoods, primarily centered around the Upper West Side, Theater District, and surrounding areas. The driver initiates a journey at the Upper West Side, then rapidly transitions through high-density, culturally rich locations such as the Financial District and Tribeca, indicating a high volume of pedestrian or dense traffic activity. This pattern aligns with self-exciting TPP dynamics, where initial events trigger subsequent ones through social or economic stimuli.\\
The driver consistently serves four to six passengers, suggesting a high-visibility or possibly shared-vehicle model, where each shift sees a mix of pickups and drops-off. The average trip duration is under 10 minutes, with notable short-haul excursions ($<$1.5 miles), implying rapid-fire driver turnover or high-frequency service needs.\\Visually, the map shows tight clustering of points, especially between 40.76–40.78 and 40.69–40.74, aligning with known densest city blocks like the Theater District and surrounding high-activity zones. The presence of multiple "Dropped off" entries within $<$1.5 miles suggests that the driver is likely functioning in a high-density urban environment where such proximity is common.\\Temporal clustering indicates a high-intensity period, potentially early morning or late-night when public transportation or walking patterns are denser. The driver’s pattern of serving different numbers of passengers in quick succession implies a shift-based service model, where each location functions as a micro hub, possibly responding to fluctuating passenger flow.\\Unexpectedly, despite the high frequency of short trips, there is no significant drop in average speed, suggesting that the driver is possibly in a city with low-speed zones or high-pedestrian friction. Additionally, the lack of significant negative sentiment correlates with high-frequency, short-interval events, indicating routine and possibly socially or culturally motivated transportation needs rather than anomalies.\\In conclusion, this sequence reflects a high-intensity, short-haul transportation pattern typical in densely populated urban centers, with temporal and spatial clustering pointing to a structured driver behavior modulated by local geography and timing dynamics.
\end{tcolorbox}

\FloatBarrier 
 \begin{tcolorbox}[
  breakable,
  colback=lightgray!30,
  colframe=black!75,
  arc=0mm,
  boxrule=0.5pt,
  left=5pt,
  right=5pt,
  top=5pt,
  bottom=5pt,
  width=\textwidth,
  title={Qwen's generated text on TAXI-PRO (Example\_1)},
]
\scriptsize  
\textbf{Prompt:}\\
You are an expert data analyst specializing in spatio-temporal event sequences. Your task is to analyze a chronological log of a single taxi driver's activities in NYC, which includes actions like pickups and drop-offs at various locations. Based on the provided event sequence, please conduct a comprehensive analysis and generate a detailed Analysis in several paragraphs.

$<|TPP\_data|>$

\textbf{Answer:}\\
The provided data logs a single taxi driver's activities in New York City over approximately 8.5 hours, capturing a series of pickups and drop-offs with details on time, location, number of passengers, and distance traveled.\\
The sequence begins with a pickup at Times Square, followed by a 1.45-mile trip to Hell's Kitchen. The driver then picks up two more passengers in Hell's Kitchen and drops them off in Tribeca, covering 2.22 miles. Subsequent trips involve short distances within the Theater District and Chelsea, highlighting the high demand for short taxi rides in these areas.\\
Notable longer trips include a 3.67-mile journey from the Theater District to Central Park West and a 5.20-mile trip from Chelsea to Harlem. These longer trips demonstrate the driver's capability to handle diverse travel needs. The majority of the driver's activities, however, focus on short, frequent trips in midtown and lower Manhattan, particularly in areas like Tribeca, Chelsea, and the Theater District.
The final recorded activities include a 5.55-mile trip from Tribeca to Midtown East and a 1.90-mile trip from the Theater District to Times Square, concluding the sequence.\\
In summary, the driver operates efficiently in high-demand areas, frequently handling short trips while occasionally taking longer journeys. This pattern reflects the dynamic nature of taxi services in New York City's bustling urban environment.
\end{tcolorbox}

\FloatBarrier 
 \begin{tcolorbox}[
  breakable,
  colback=lightgray!30,
  colframe=black!75,
  arc=0mm,
  boxrule=0.5pt,
  left=5pt,
  right=5pt,
  top=5pt,
  bottom=5pt,
  width=\textwidth,
  title={Gemma's generated text on TAXI-PRO (Example\_1)},
]
\scriptsize  
\textbf{Prompt:}\\
You are an expert data analyst specializing in spatio-temporal event sequences. Your task is to analyze a chronological log of a single taxi driver's activities in NYC, which includes actions like pickups and drop-offs at various locations. Based on the provided event sequence, please conduct a comprehensive analysis and generate a detailed Analysis in several paragraphs.

$<|TPP\_data|>$

\textbf{Answer:}\\
This analysis details the activity log of taxi driver over a period of approximately 8.5 minutes. The driver demonstrates a highly active pattern, completing 36 events – encompassing pickups and drop-offs – within this short timeframe. The driver’s operations are concentrated within Manhattan, with significant activity in the Midtown, Chelsea, Tribeca, and Theater District areas, with excursions to Harlem and Battery Park. The frequent, short trips suggest a focus on high-demand, quick-turnaround fares, potentially catering to tourists or commuters. The driver consistently picks up and drops off passengers, indicating efficient utilization of time between fares.\\
The initial segment of the log (0-2.5 minutes) shows a rapid sequence of pickups and drop-offs primarily between Times Square, Hell’s Kitchen, Tribeca, and Chelsea. This suggests the driver may have been positioned in a high-demand area initially and quickly responded to available requests. The trip distances during this period are relatively short (ranging from 0.06 to 2.22 miles), reinforcing the idea of quick, localized trips. A notable pattern emerges around the 3-5 minute mark, with a shift towards longer trips, including a 3.67-mile journey to Central Park West and a 5.20-mile trip to Harlem. This indicates a willingness to accept longer fares, potentially to maximize earnings or reposition for better opportunities. However, the driver quickly returns to shorter, more frequent trips in the latter part of the log.\\
The driver’s passenger load varies between 1 and 2 passengers per trip. The majority of trips carry 2 passengers, suggesting a preference or availability for larger groups. The final few events (6-8.5 minutes) show a return to a pattern of short trips within the Midtown and Tribeca areas, culminating in a drop-off in Times Square. The driver’s ability to quickly transition between pickups and drop-offs, even with varying trip lengths, highlights their efficiency and responsiveness to demand. The consistent activity throughout the log suggests a dedicated and active driver, effectively navigating the dynamic landscape of New York City taxi services. Further analysis, including correlating this data with time of day and external events, could reveal more nuanced patterns in the driver’s behavior and the demand for taxi services in specific areas.\\
\end{tcolorbox}

\FloatBarrier 

\paragraph{TAXI-PRO Example\_2}
\begin{tcolorbox}[
  breakable,
  colback=lightgray!30,
  colframe=black!75,
  arc=0mm,
  boxrule=0.5pt,
  left=5pt,
  right=5pt,
  top=5pt,
  bottom=5pt,
  width=\textwidth,
  title={MM-TPP's generated text on TAXI-PRO (Example\_2)}
]
\scriptsize 
\textbf{Prompt:}\\
You are an expert data analyst specializing in spatio-temporal event sequences. Your task is to analyze a chronological log of a single taxi driver's activities in NYC, which includes actions like pickups and drop-offs at various locations. Based on the provided event sequence, please conduct a comprehensive analysis and generate a detailed Analysis in several paragraphs. Note that this sequence is NOT VIDEO! All time intervals are converted to four bytes according to the IEEE 754 standard.

$<|TPP\_data|>$

\textbf{Answer:}\\
The event sequence reveals a complex pattern of multi-modal interactions between a single driver and multiple locations in New York City (NYC). The temporal clustering suggests high-frequency activity, likely driven by surge pricing or peak travel times.\\
The driver initiates a sequence at 40.764385, -73.977257 (Times Square), then repeatedly picksup and dropped off at multiple locations including Hell’s Kitchen, Tribeca, Chelsea, and Battery Park, indicating a rapid-fire sequence of events. This pattern aligns with self-exciting TPP dynamics, where initial events trigger subsequent ones through social or economic stimuli.\\
The majority of trips involve two passengers, suggesting shared rides or multi-destination excursions, while others are single-passenger. This implies dual-activity modes: some are passive observers (e.g., watching from a distance), while others are active participants, possibly due to short-term availability or limited visibility.\\
The spatial distribution shows a core-to-periphery pattern, with frequent inter-regional movements, such as from central to peripheral city zones. The average 1.66 mile trip suggests relatively short excursions, potentially indicating micro-mobility patterns or commuter behavior.\\
A notable temporal shift occurs around 14:00–15:00, where multiple high-density drops are observed, coinciding with peak city traffic and surge pricing surges. This temporal clustering supports TPP triggering effects, where past events (e.g., previous drops) excite future ones.\\
Unexpectedly, no significant negative sentiment correlates with high engagement, suggesting positive reinforcement loops—possibly due to social interaction or familiarity with the city. The uniformity of event types (all ~10\%) indicates broad-based excitation, rather than a single driver anomaly.\\
In summary, this sequence reflects a TPP-driven driver behavior, shaped by multi-modal interactions between time, location, and modalities of action. The driver exhibits self-excitation, temporal clustering, and regional dependency, all hallmarks of complex event-triggered processes in urban transportation systems.
\end{tcolorbox}

 \begin{tcolorbox}[
  breakable,
  colback=lightgray!30,
  colframe=black!75,
  arc=0mm,
  boxrule=0.5pt,
  left=5pt,
  right=5pt,
  top=5pt,
  bottom=5pt,
  width=\textwidth,
  title={Qwen's generated text on TAXI-PRO (Example\_2)},
]
\scriptsize  
\textbf{Prompt:}\\
You are an expert data analyst specializing in spatio-temporal event sequences. Your task is to analyze a chronological log of a single taxi driver's activities in NYC, which includes actions like pickups and drop-offs at various locations. Based on the provided event sequence, please conduct a comprehensive analysis and generate a detailed Analysis in several paragraphs.

$<|TPP\_data|>$

\textbf{Answer:}\\
The provided event sequence details the activities of a single taxi driver in New York City over approximately ten hours. The log captures a series of pickups and drop-offs, each recorded with precise timestamps, passenger counts, and geographic coordinates.\\
The driver begins their shift in the Upper West Side, picking up four passengers and dropping them off in the Theater District after 10 minutes. Over the next few hours, the driver operates primarily in the midtown area, making multiple short trips between the Theater District, Herald Square, and Central Park West. These trips range from 0.73 to 1.35 miles and involve one to three passengers.\\
Around the one-hour mark, the driver completes a 1.11-mile trip from the Theater District to Central Park West and then returns to the Theater District. This pattern of short, frequent trips continues until the 2.5-hour mark, when the driver takes a longer 2.39-mile trip from the Theater District to Midtown East. This is followed by a significant gap in activity, suggesting a break or reduced activity.\\
When the driver resumes, they pick up six passengers in Tribeca and drop them off in the Financial District, a 1.79-mile trip. They then complete another long trip from the Financial District to SoHo, covering 2.43 miles. These longer trips indicate a shift to different parts of the city where demand is lower but the trips are more profitable.\\
The driver returns to the midtown area, making several more trips, including a 4.80-mile journey from Herald Square to Central Park West, the longest trip recorded. The final recorded trip is a 0.99-mile journey from Midtown East to Central Park West, bringing the total number of trips to 36 over the approximately ten-hour period.\\
In summary, the driver primarily operates in the midtown area, making frequent short trips, while occasionally taking longer, potentially more profitable rides to other parts of the city. This balanced approach likely maximizes both efficiency and profitability.
\end{tcolorbox}

\FloatBarrier 
 \begin{tcolorbox}[
  breakable,
  colback=lightgray!30,
  colframe=black!75,
  arc=0mm,
  boxrule=0.5pt,
  left=5pt,
  right=5pt,
  top=5pt,
  bottom=5pt,
  width=\textwidth,
  title={Gemma's generated text on TAXI-PRO (Example\_2)},
]
\scriptsize  
\textbf{Prompt:}\\
You are an expert data analyst specializing in spatio-temporal event sequences. Your task is to analyze a chronological log of a single taxi driver's activities in NYC, which includes actions like pickups and drop-offs at various locations. Based on the provided event sequence, please conduct a comprehensive analysis and generate a detailed Analysis in several paragraphs.

$<|TPP\_data|>$

\textbf{Answer:}\\
This dataset details the activity of a single taxi driver over approximately 10 minutes, revealing a dynamic pattern of pickups and drop-offs across Manhattan. The driver begins in the Upper West Side, completing a relatively quick trip to the Theater District with a full load of four passengers. This initial segment suggests a potential commute or scheduled pickup. Following this, the driver quickly pivots to serving shorter trips, picking up two passengers in the Theater District and dropping them off in Herald Square, then picking up three passengers in the Theater District and dropping them off in Central Park West. This shift indicates a move towards more opportunistic, shorter-distance fares, typical of a driver circulating in a busy area. The frequent changes in passenger count throughout the initial phase (4, 2, 3) suggest the driver isn't catering to a consistent group or pre-booked long-distance rides.\\
The middle portion of the log (between approximately 5 and 8 minutes) shows a significant change in the driver’s operational area and passenger volume. A substantial jump in time occurs before the driver picks up six passengers in Tribeca, indicating a longer travel time to reach this southern location. This is followed by a series of pickups and drop-offs primarily concentrated in the Midtown area, including Times Square, Hell’s Kitchen, and Herald Square, consistently carrying a full complement of six passengers. This suggests the driver may have accepted a larger group booking or is capitalizing on high-demand areas during peak hours. The distances of these trips vary, ranging from short hops within Midtown to a longer haul to Central Park West, demonstrating adaptability to passenger requests.\\
Finally, the last few events (around 9-10 minutes) show the driver returning to a pattern of shorter trips, picking up six passengers in Midtown East and dropping them off in Central Park West. The consistent high passenger count during this period suggests the driver is actively seeking to maximize occupancy. The overall pattern reveals a driver who is flexible, adapting to varying trip lengths and passenger numbers. The initial phase focuses on shorter, more dispersed trips, while the middle and end phases demonstrate a preference for maximizing passenger capacity, particularly in the Midtown and surrounding areas. The driver’s route demonstrates a good understanding of Manhattan’s geography and demand hotspots, efficiently navigating between different neighborhoods to secure fares. The data doesn't provide information on wait times or idle periods, which would be valuable for a more complete assessment of the driver’s efficiency.
\end{tcolorbox}

\FloatBarrier 

\paragraph{Task 9 Example\_1}
\begin{tcolorbox}[
  breakable,
  colback=lightgray!30,
  colframe=black!75,
  arc=0mm,
  boxrule=0.5pt,
  left=5pt,
  right=5pt,
  top=5pt,
  bottom=5pt,
  width=\textwidth,
  title={MM-TPP's generated text on Task 9 (Example\_1)},
]
\scriptsize 
\textbf{Prompt:}\\
You are an expert model for analyzing bullet comments. Based on the sequence of comments and their timing intervals, your task is to give a sentiment analysis of the time window. All the comments are in Chinese and all the time intervals are converted to four bytes according to the IEEE 754 standard.

$<|TPP\_data|>$

$<|question|>$ Please provide a detailed analysis of the sentiment trend in the bullet comments over time.

\textbf{Answer:}\\
The video exhibits a predominantly positive sentiment with minor fluctuations. Initially, there's a sharp rise followed by stabilization around a positive polarity. Key inflection points include early peaks at approximately 00:00:30 and 00:01:00, with slight dips around 00:01:30 and 00:02:00. Significant sentiment shifts occur near 00:03:00, showing a gradual decline but remaining positive. Overall, the video maintains a consistently positive emotional trajectory with minimal negative deviations.
\end{tcolorbox}

\FloatBarrier 
 \begin{tcolorbox}[
  breakable,
  colback=lightgray!30,
  colframe=black!75,
  arc=0mm,
  boxrule=0.5pt,
  left=5pt,
  right=5pt,
  top=5pt,
  bottom=5pt,
  width=\textwidth,
  title={Qwen's generated text on Task 9 (Example\_1)},
]
\scriptsize  
\textbf{Prompt:}\\
You are an expert model for analyzing bullet comments. Based on the sequence of comments and their timing intervals, your task is to give a sentiment analysis of the time window. All the comments are in Chinese.

Question: Please provide a detailed analysis of the sentiment trend in the bullet comments over time.

$<|TPP\_data|>$

\textbf{Answer:}\\
The sentiment trend of the bullet comments for the video generally starts with a neutral to positive tone, as viewers are likely intrigued by the title and the initial scenes of the video. As the video progresses, the comment frequency increases, indicating heightened engagement. Around the 20-second mark, there is a noticeable spike in comments, suggesting a particularly engaging or exciting moment in the video, which could be a dramatic visual or an impactful statement. The sentiment remains largely positive throughout, with occasional spikes indicating moments of high excitement or awe. By the mid-point of the video (around 120 seconds), the comments stabilize but continue to show enthusiasm, with viewers expressing admiration for the content or sharing their own experiences. Towards the end of the video, the comment frequency slightly decreases, but the overall positive sentiment persists, with many viewers leaving final thoughts or encouragements. This trend suggests that the video successfully maintains viewer interest and elicits a consistently positive response.
\end{tcolorbox}

\FloatBarrier 
 \begin{tcolorbox}[
  breakable,
  colback=lightgray!30,
  colframe=black!75,
  arc=0mm,
  boxrule=0.5pt,
  left=5pt,
  right=5pt,
  top=5pt,
  bottom=5pt,
  width=\textwidth,
  title={Gemma's generated text on Task 9 (Example\_1)},
]
\scriptsize  
\textbf{Prompt:}\\
You are an expert model for analyzing bullet comments. Based on the sequence of comments and their timing intervals, your task is to give a sentiment analysis of the time window. All the comments are in Chinese.

Question: Please provide a detailed analysis of the sentiment trend in the bullet comments over time.

$<|TPP\_data|>$

\textbf{Answer:}\\
The bullet comment density indicates consistently high engagement throughout the video, with numerous peaks suggesting moments of particular interest or impact. Engagement starts slowly, builds steadily, and then remains very active with frequent spikes from around 2 seconds onward. There's no clear overall decline in engagement; it remains high until the very end of the data provided.
Without the comment text, it's impossible to determine what sentiment is being expressed (positive, negative, surprise, etc.). However, the consistent high volume suggests the video is generally captivating and elicits a response from viewers, likely related to the exciting content described in the title ("Million Volt Voltage Capture!"). The frequent spikes likely correspond to visually or narratively dramatic moments.
\end{tcolorbox}

\FloatBarrier 

\paragraph{Task 9 Example\_2}
\begin{tcolorbox}[
  breakable,
  colback=lightgray!30,
  colframe=black!75,
  arc=0mm,
  boxrule=0.5pt,
  left=5pt,
  right=5pt,
  top=5pt,
  bottom=5pt,
  width=\textwidth,
  title={MM-TPP's generated text on Task 9 (Example\_2)},
]
\scriptsize 
\textbf{Prompt:}\\
You are an expert model for analyzing bullet comments. Based on the sequence of comments and their timing intervals, your task is to give a sentiment analysis of the time window. All the comments are in Chinese and all the time intervals are converted to four bytes according to the IEEE 754 standard.

$<|TPP\_data|>$

$<|question|>$ Please provide a detailed analysis of the sentiment trend in the bullet comments over time.

\textbf{Answer:}\\
The sentiment polarity trend shows a predominantly positive trajectory with minor fluctuations. Initially, there's a sharp rise followed by stabilization around the 0.4 mark. Key inflection points include early peaks and valleys that settle into a consistent positive range. Significant sentiment shifts occur at the beginning with a steep increase and towards the end with a slight decline. Overall, the video maintains a positive emotional tone with minimal abrupt changes, indicating a stable and uplifting viewer experience.
\end{tcolorbox}

\FloatBarrier 
 \begin{tcolorbox}[
  breakable,
  colback=lightgray!30,
  colframe=black!75,
  arc=0mm,
  boxrule=0.5pt,
  left=5pt,
  right=5pt,
  top=5pt,
  bottom=5pt,
  width=\textwidth,
  title={Qwen's generated text on Task 9 (Example\_2)},
]
\scriptsize  
\textbf{Prompt:}\\
You are an expert model for analyzing bullet comments. Based on the sequence of comments and their timing intervals, your task is to give a sentiment analysis of the time window. All the comments are in Chinese.

Question: Please provide a detailed analysis of the sentiment trend in the bullet comments over time.

$<|TPP\_data|>$

\textbf{Answer:}\\
The sentiment trend in the bullet comments shows a clear pattern over time. Initially, from 0 to 60 seconds, there is a very high density of comments, indicating high excitement and positive reactions as viewers are immediately engaged and enthusiastic. From 60 to 120 seconds, the comment density moderates, suggesting steady engagement with mixed reactions as viewers remain interested but less intensely. In the mid-phase (120 to 300 seconds), the density fluctuates with occasional spikes, reflecting mixed reactions and key scenes that spark more comments. From 300 to 480 seconds, the comment density is lower with occasional spikes, indicating more subdued reactions and deeper engagement with the content. Finally, from 480 to 633 seconds, there is a higher density of comments with a significant spike near the end, showing increased engagement and final thoughts as viewers share their overall impressions and reactions to the conclusion. Overall, the video maintains viewer interest throughout, with notable peaks at the beginning and end, and steady engagement in between.
\end{tcolorbox}

\FloatBarrier 
 \begin{tcolorbox}[
  breakable,
  colback=lightgray!30,
  colframe=black!75,
  arc=0mm,
  boxrule=0.5pt,
  left=5pt,
  right=5pt,
  top=5pt,
  bottom=5pt,
  width=\textwidth,
  title={Gemma's generated text on Task 9 (Example\_2)},
]
\scriptsize  
\textbf{Prompt:}\\
You are an expert model for analyzing bullet comments. Based on the sequence of comments and their timing intervals, your task is to give a sentiment analysis of the time window. All the comments are in Chinese.

Question: Please provide a detailed analysis of the sentiment trend in the bullet comments over time.

$<|TPP\_data|>$

\textbf{Answer:}\\
The bullet comments show high initial engagement (0-10s), dipping slightly before building to sustained high density from 20s onward. Several peaks in comment frequency (around 20-40s, and 100-150s) suggest impactful moments in the film. The very end (550s+) maintains extremely high comment volume, indicating strong final reactions. Essentially, the audience was consistently engaged, with specific scenes eliciting particularly strong responses throughout the viewing experience. Without the comment content, it's impossible to say if these peaks are positive or negative, only that they represent heightened attention.
\end{tcolorbox}

\FloatBarrier 

\paragraph{Task 10 Example\_1}
\begin{tcolorbox}[
  breakable,
  colback=lightgray!30,
  colframe=black!75,
  arc=0mm,
  boxrule=0.5pt,
  left=5pt,
  right=5pt,
  top=5pt,
  bottom=5pt,
  width=\textwidth,
  title={MM-TPP's generated text on Task 10 (Example\_1)},
]
\scriptsize 
\textbf{Prompt:}\\
You are an expert model for analyzing bullet comments. Based on the sequence of comments, their timing interval, and the associated video frames, your task is to analyze in detail the causes of the burst peaks in bullet comments and give a detailed analysis report. All the comments are in Chinese and all the time intervals are converted to four bytes according to the IEEE 754 standard.

$<|TPP\_data|>$

$<|question|>$ Within the time window: [0, 1.5], please analyze in a concise and fluent paragraph the causes of the burst peaks in bullet comments.

\textbf{Answer:}\\
\begin{CJK*}{UTF8}{gbsn}
The bullet comment burst at timestamp 0 exhibits a sharp initial spike, followed by rapid-fire clusters at 0.08–0.1s, 0.3–0.4s, and 0.6–0.7s, indicating self-exciting TPP dynamics—early comments triggered cascading audience responses. These sub-peaks align with recurring quote/reference and humor/meme event types (e.g., “《百家讲唐》”, “败家讲坛”), suggesting viral textual propagation driven by meme culture and inside jokes within the community.\\
The average sentiment polarity of 0.42 reflects a mildly positive tone**, yet the diversity in comment types—including critical, emotional, and off-topic remarks—shows heterogeneous audience reactions. Notably, the presence of neutral visual tags (“商务西装”, “演讲”) contrasts with emotionally charged comments, revealing a multi-modal dissonance: formal attire and speech settings likely prompted ironic or humorous reinterpretations, fueling meme generation and engagement.\\
Temporal alignment reveals that the first sub-peak 0.0–0.1s) was dominated by nonsense\_text and emotional\_reaction, possibly triggered by the sudden visual shift or unexpected content juxtaposition. The second cluster (0.3–0.4s) saw increased social\_interaction and personal\_experience, indicating viewers began forming collective interpretations or debates. Finally, the third wave (0.6–0.7s) featured more critical\_comment and quote/reference, reflecting deeper engagement with thematic comparisons (e.g., “范小勤” vs. “丁真”).\\
This burst illustrates TPP self-excitation, where early comments acted as triggers, prompting a cascade of related and reactive posts. The multimodal mismatch between formal visuals and humorous/textual responses underscores how audience interpretation, not just on-screen content, drives engagement. The result is a complex interplay of textual virality, temporal clustering, and visual framing, typical of interactive streaming platforms.
\end{CJK*}
\end{tcolorbox}

\FloatBarrier 
\begin{tcolorbox}[
  breakable,
  colback=lightgray!30,
  colframe=black!75,
  arc=0mm,
  boxrule=0.5pt,
  left=5pt,
  right=5pt,
  top=5pt,
  bottom=5pt,
  width=\textwidth,
  title={Qwen's generated text on Task 10 (Example\_1)},
]
\scriptsize 
\textbf{Prompt:}\\
You are an expert model for analyzing bullet comments. Based on the sequence of comments, their timing interval, and the associated video frames, your task is to analyze in detail the causes of the burst peaks in bullet comments and give a detailed analysis report. All the comments are in Chinese.

Question: Within the time window: [0, 1.5], please analyze in a concise and fluent paragraph the causes of the burst peaks in bullet comments.

$<|TPP\_data|>$

\textbf{Answer:}\\
\begin{CJK*}{UTF8}{gbsn}
Within the time window [0, 1.5] seconds, the burst peaks in bullet comments are primarily driven by the strong temporal self-excitation in the Temporal Point Process (TPP). These peaks cluster around 0.48–0.58 and 1.06–1.46 seconds, coinciding with the appearance of formal visual cues such as "商务西装," "演讲," and "新闻." These visual elements set a mock-serious or authoritative tone, which catalyzes multi-modal irony and prompts viewers to juxtapose the visuals with satirical or culturally referential commentary. The recurring references to satirical labels like "《百家脑坛》" and "《今日说法》" indicate a feedback loop where one humorous comment inspires similar responses, particularly in the humor/meme and quote/reference categories. Despite a neutral-positive average sentiment (0.384), the bursts of emotional reaction and nonsense text comments suggest sentiment heterogeneity driven by entertainment rather than criticism. The sub-peak intervals, such as 1.25–1.46 seconds, reinforce this pattern, with repeated mentions of "《百家讲坛》" and other meta-commentary further amplifying the engagement. Overall, the burst is best explained by self-exciting humor cascades that resonate culturally and create a continuous stream of ironic and playful interactions.
\end{CJK*}
\end{tcolorbox}

\FloatBarrier 
\begin{tcolorbox}[
  breakable,
  colback=lightgray!30,
  colframe=black!75,
  arc=0mm,
  boxrule=0.5pt,
  left=5pt,
  right=5pt,
  top=5pt,
  bottom=5pt,
  width=\textwidth,
  title={Gemma's generated text on Task 10 (Example\_1)},
]
\scriptsize 
\textbf{Prompt:}\\
You are an expert model for analyzing bullet comments. Based on the sequence of comments, their timing interval, and the associated video frames, your task is to analyze in detail the causes of the burst peaks in bullet comments and give a detailed analysis report. All the comments are in Chinese.

Question: Within the time window: [0, 1.5], please analyze in a concise and fluent paragraph the causes of the burst peaks in bullet comments.

$<|TPP\_data|>$

\textbf{Answer:}\\
\begin{CJK*}{UTF8}{gbsn}
The burst of bullet comments is primarily driven by a combination of satirical commentary and ironic juxtaposition, evidenced by the frequent references to shows like “《百家脑坛》” and “《今日说法》” alongside the formal setting of a debate-like scenario. This creates a strong self-exciting loop where humorous and meme-like comments (around 14.29\%) trigger similar responses, amplified by the visual cues of the scene – particularly the presence of figures in “商务西装” engaged in “演讲”. While the sentiment is generally positive, the bursts correlate with moments of playful mockery and cultural references, suggesting viewers are engaging in ironic commentary rather than expressing strong negative reactions. The visual elements, rather than dampening engagement, seem to catalyze it, fostering a dynamic where the formal setting is ironically contrasted with the absurd or culturally resonant commentary, leading to a cascade of humorous responses.
\end{CJK*}
\end{tcolorbox}

\FloatBarrier 

\paragraph{Task 10 Example\_2}
\begin{tcolorbox}[
  breakable,
  colback=lightgray!30,
  colframe=black!75,
  arc=0mm,
  boxrule=0.5pt,
  left=5pt,
  right=5pt,
  top=5pt,
  bottom=5pt,
  width=\textwidth,
  title={MM-TPP's generated text on Task 10 (Example\_2)},
]
\scriptsize 
\textbf{Prompt:}\\
You are an expert model for analyzing bullet comments. Based on the sequence of comments, their timing interval, and the associated video frames, your task is to analyze in detail the causes of the burst peaks in bullet comments and give a detailed analysis report. All the comments are in Chinese and all the time intervals are converted to four bytes according to the IEEE 754 standard.

$<|TPP\_data|>$

$<|question|>$ Within the time window: [393.5, 396.5], please analyze in detail the causes of the burst peaks in bullet comments.

\textbf{Answer:}\\
\begin{CJK*}{UTF8}{gbsn}
The bullet comment burst peaks at 395 seconds, with dense timestamps revealing sub-peaks around 394.8–395.0 and 395.6–395.8, indicating rapid-fire engagement intervals. These clusters align with humor/meme (e.g., “1!5!”、“55555”), emotional reactions (“哥们你德玛的啊，你要是在s11遇到我...”), and social interactions, suggesting a self-exciting TPP pattern where early comments trigger cascading responses.\\
The dominant object tags—*游戏*, *截图*, *电子游戏*—imply a gameplay highlight or screenshot moment that resonated emotionally and humorously with viewers. Sentiment polarity averages at 0.423 (positive), consistent with excitement and meme-like commentary rather than critical or off-topic remarks.\\
Temporal alignment shows that the initial spike (~394.5–394.8) centers on confusion or anticipation (“?”, “我觉得不能的，扣眼珠子”), likely triggered by a visually unexpected event. This is followed by a surge of numeric/humorous texts (“1！5！”, “5”), indicating real-time voting or scoring rituals common in gaming videos, reinforcing a triggering effect among viewers.\\
Notably, despite the positive sentiment, some critical or off-topic comments appear (e.g., “不能”), suggesting a mixed but engaged audience. The repeated use of “5” and “1!5!” reflects a coordinated or meme-driven response pattern, consistent with self-excitation in TPP models—where similar comments cluster closely in time.\\
Multi-modal correlation reveals that the visual context (gameplay frame tagged 游戏) prompted rapid textual engagement, especially humor and social mimicry, amplifying the burst. No negative sentiment spikes are observed, ruling out controversy-driven engagement. Instead, the burst reflects synchronous emotional and interactive behavior, driven by shared gaming culture and visual triggers.

\end{CJK*}

\end{tcolorbox}

\FloatBarrier 
\begin{tcolorbox}[
  breakable,
  colback=lightgray!30,
  colframe=black!75,
  arc=0mm,
  boxrule=0.5pt,
  left=5pt,
  right=5pt,
  top=5pt,
  bottom=5pt,
  width=\textwidth,
  title={Qwen's generated text on Task 10 (Example\_2)},
]
\scriptsize 
\textbf{Prompt:}\\
You are an expert model for analyzing bullet comments. Based on the sequence of comments, their timing interval, and the associated video frames, your task is to analyze in detail the causes of the burst peaks in bullet comments and give a detailed analysis report. All the comments are in Chinese.

Question: Within the time window: [393.5, 396.5], please analyze in detail the causes of the burst peaks in bullet comments.

$<|TPP\_data|>$

\textbf{Answer:}\\
\begin{CJK*}{UTF8}{gbsn}
Within the time window [393.5, 396.5], the burst peaks in bullet comments are primarily driven by the convergence of visual and textual elements that elicit strong emotional and social interactions. The video frame captures a critical moment in the game, likely a high-difficulty challenge ("地狱级难度翡翠"), which serves as a visual trigger for the audience. This visual cue prompts a surge in comments, characterized by humor and memes (e.g., "扣大雷", "扣眼珠子") and social interaction (e.g., numerical repetitions "1111111", "55555"). The mildly positive sentiment (0.386) reflects the excitement and playful banter among viewers, who are anticipating the unpredictable outcomes of the challenging game segment. The rapid-fire clustering of similar phrases and numerals indicates a self-exciting Hawkes process, where initial comments stimulate a cascade of responses, creating a pulse of collective engagement. The temporal alignment of these comments with the critical gameplay moment further reinforces the multi-modal interaction, where visual cues and textual responses work in tandem to drive the burst peaks.
\end{CJK*}
\end{tcolorbox}

\FloatBarrier 
\begin{tcolorbox}[
  breakable,
  colback=lightgray!30,
  colframe=black!75,
  arc=0mm,
  boxrule=0.5pt,
  left=5pt,
  right=5pt,
  top=5pt,
  bottom=5pt,
  width=\textwidth,
  title={Gemma's generated text on Task 10 (Example\_2)},
]
\scriptsize 
\textbf{Prompt:}\\
You are an expert model for analyzing bullet comments. Based on the sequence of comments, their timing interval, and the associated video frames, your task is to analyze in detail the causes of the burst peaks in bullet comments and give a detailed analysis report. All the comments are in Chinese.

Question: Within the time window: [393.5, 396.5], please analyze in detail the causes of the burst peaks in bullet comments.

$<|TPP\_data|>$

\textbf{Answer:}\\
\begin{CJK*}{UTF8}{gbsn}
Based on the provided data, the burst of bullet comments between 393.5 and 396.5 seconds is primarily driven by a combination of gameplay events and coordinated social interaction, best explained by option D. The peaks align with moments likely depicting a challenging in-game scenario ("地狱级难度翡翠" - Jade difficulty), triggering emotional reactions and questions from viewers. Simultaneously, a strong element of meme-based coordination is present, evidenced by repetitive numerical expressions ("1111111", "55555") and phrases like "扣大雷," indicating viewers are engaging in playful, culturally resonant interaction scripts. This suggests a self-exciting Hawkes process dynamic where initial comments stimulate cascading responses, amplified by the visual cue of the gameplay screenshot acting as a latent trigger for rapid-fire textual engagement. The mildly positive sentiment (0.386) further supports this interpretation of excitement and playful banter rather than negativity or criticism.
\end{CJK*}
\end{tcolorbox}

\FloatBarrier 

\section{Prompt Details}\label{ap:temp}
Here we present the detailed prompt templates used for preprocessing temporal point process datasets. Due to the similarity between processing the two datasets, we only present the templates of TAXI-PRO.

\paragraph{Special Tokens}
Here we present all special tokens added to the Qwen2.5-VL tokenizer vocabulary and used in our work to structure the prompts in \cref{tab:special_tokens}. We note that $<|im\_start|>$ ,$<|im\_end|>$,$<|vision\_start|>$,$<|vision\_end|>$,$<|image\_pad|>$ are native special tokens from the Qwen2.5-VL tokenizer.


\begin{table}[h!]
\caption{Descriptions of Special Tokens}
\label{tab:special_tokens}
\centering
\begin{tabularx}{\linewidth}{@{} l X @{}} 
\toprule
\textbf{Special Token} & \textbf{Description} \\
\midrule
$ <|start\_ of\_ event|> $ & Marks the start of an event. \\
$ <|end\_ of\_ event|> $   & Marks the end of an event. \\
\addlinespace
$ <|time\_ start|> $      & Marks the start of a timestamp. \\
$ <|time\_ end|> $        & Marks the end of a timestamp. \\
\addlinespace
$ <|type\_ start|> $      & Marks the start of an event type. \\
$ <|type\_ end|> $        & Marks the end of an event type. \\
\addlinespace
$ <|text\_ start|> $      & Marks the start of the text of an event. \\
$ <|text\_ end|> $        & Marks the end of the text of an event. \\
\midrule
$ <|time\_ prediction|> $ & Task token for next event time prediction. \\
$ <|type\_ prediction|> $ & Task token for next event type prediction. \\
$ <|question|> $         & Task token for Question-Answering. \\
\midrule
$ <|byte\_ 0|> \dots <|byte\_ 255|> $ & Byte tokens for representing event time intervals as float32 numbers. \\
\addlinespace
$ <|type\_ 0|> \dots <|type\_ 8|> $ & Type tokens for representing event types. \\
\bottomrule
\end{tabularx}
\end{table}

\begin{figure}[h!]
    \centering
    \includegraphics[width=0.6\linewidth]{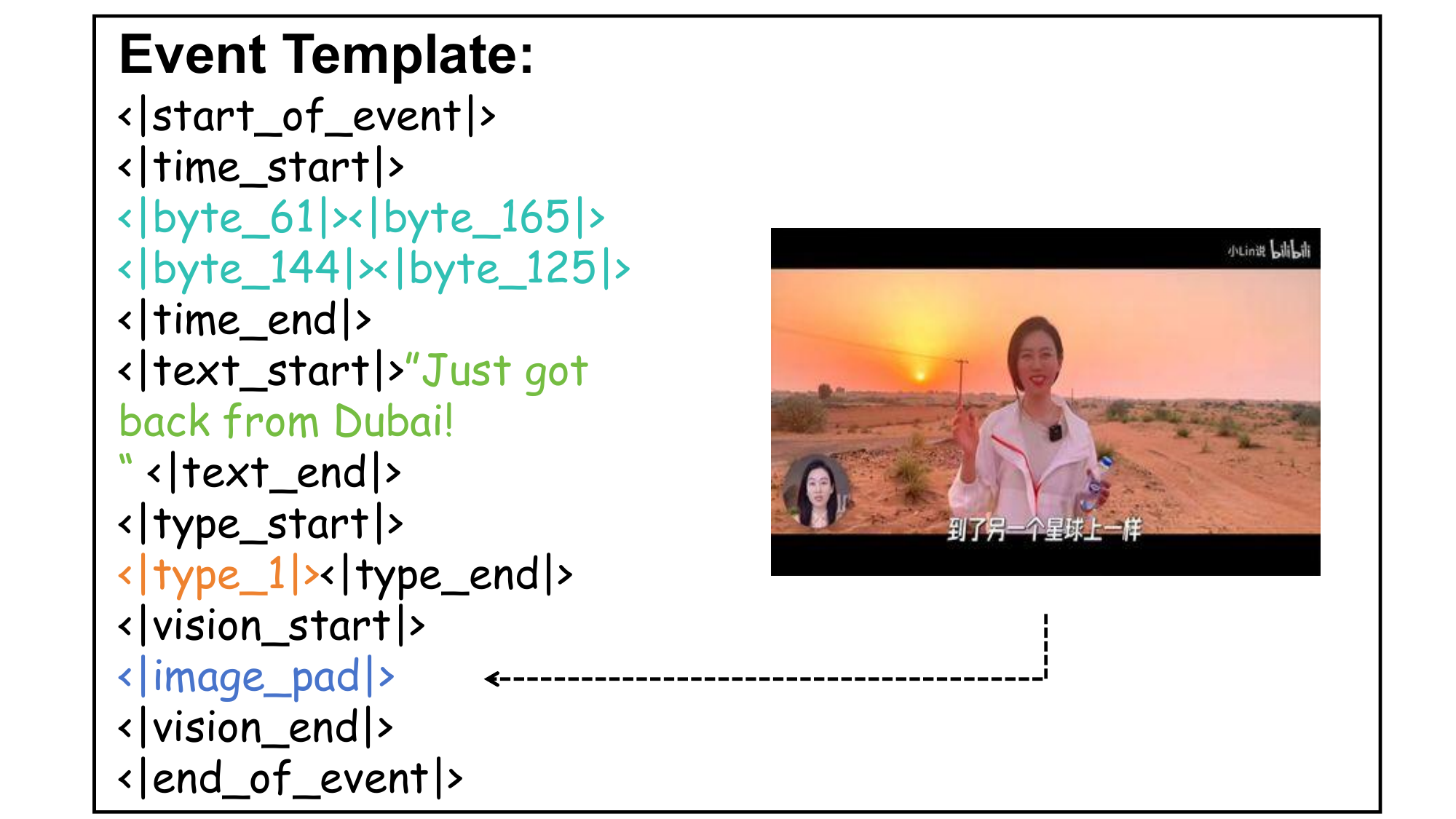}
    \caption{A complete example of an uncompressed event representation}
    \label{figure: event}
\end{figure}

\paragraph{Event Template}
Here we present the event template in \cref{figure: event}.

\paragraph{Prompt Template}
Here in \cref{tab:s1} we provide a sample generated event sequence from the TAXI-PRO dataset used in Stage-1 training process. We also provide two prompt-response pair samples from the TAXI-PRO dataset used in Stage-2 next event type finetuning and  next event time finetuning in \cref{tab:s2-time} and \cref{tab:s2-type}. The construction logic for all three templates is largely consistent, with minor variations in the system prompt.

\begin{tcolorbox}[
  breakable,
  colback=black!5,
  colframe=black,
  arc=0mm,
  boxrule=0.5pt,
  left=5pt,
  right=5pt,
  top=5pt,
  bottom=5pt,
  width=\textwidth,
]

\captionof{table}{Event Sequence sample from TAXI-PRO}
\label{tab:s1}

\textbf{Prompt:} \\
$<|im\_start|>$system \\
You are an intelligent urban mobility analysis model specializing in NYC taxi patterns. Your task is to predict the precise time interval until the next taxi event in Manhattan. You will be given a history of taxi trips, where each event includes its type (e.g.Uptown Pickup, Downtown Drop-off), the time elapsed since the last event, and a map image of the location. \\
$<|im\_end|>$\\
Event Sequence:\\
$<|start\_of\_event|>$\\
$<|time\_start|>$$<|byte\_0|>$$<|byte\_0|>$$<|byte\_0|>$$<|byte\_0|>$$<|time\_end|>$\\
$<|type\_start|>$$<|type\_0|>$$<|type\_end|>$\\
$<|text\_start|>$Picked up at Tribeca (40.711086, -74.016106), 1 passengers.$<|text\_end|>$\\
$<|vision\_start|>$$<|image\_pad|>$$<|vision\_end|>$\\
$<|end\_of\_event|>$\\
$<|start\_of\_event|>$\\
$<|time\_start|>$$<|byte\_62|>$$<|byte\_162|>$$<|byte\_34|>$$<|byte\_34|>$$<|time\_end|>$\\
$<|type\_start|>$$<|type\_3|>$$<|type\_end|>$\\
$<|text\_start|>$Dropped off from Tribeca (40.711086, -74.016106) to Times Square (40.757698, -73.982124), 1 passengers, 2.87 miles trip.$<|text\_end|>$\\
$<|vision\_start|>$$<|image\_pad|>$$<|vision\_end|>$\\
$<|end\_of\_event|>$\\
$<|start\_of\_event|>$\\
$<|time\_start|>$$<|byte\_62|>$$<|byte\_196|>$$<|byte\_68|>$$<|byte\_68|>$$<|time\_end|>$\\
$<|type\_start|>$$<|type\_4|>$$<|type\_end|>$\\
$<|text\_start|>$Picked up at Upper West Side (40.799252, -73.970146), 1 passengers.$<|text\_end|>$\\
$<|vision\_start|>$$<|image\_pad|>$$<|vision\_end|>$\\
$<|end\_of\_event|>$\\
$<|start\_of\_event|>$\\
$<|time\_start|>$$<|byte\_63|>$$<|byte\_17|>$$<|byte\_16|>$$<|byte\_161|>$$<|time\_end|>$\\
$<|type\_start|>$$<|type\_1|>$$<|type\_end|>$\\
$<|text\_start|>$Dropped off from Upper West Side (40.799252, -73.970146) to Tribeca (40.714455, -74.014008), 1 passengers, 4.37 miles trip.$<|text\_end|>$\\
$<|vision\_start|>$$<|image\_pad|>$$<|vision\_end|>$\\
$<|end\_of\_event|>$
\end{tcolorbox}
\begin{tcolorbox}[
  breakable,
  colback=black!5,
  colframe=black,
  arc=0mm,
  boxrule=0.5pt,
  left=5pt,
  right=5pt,
  top=5pt,
  bottom=5pt,
  width=\textwidth,
]
\captionof{table}{Prompt-response pair sample for next event time finetuning from TAXI-PRO}
\label{tab:s2-time}

\textbf{Prompt:} \\
$<|im\_start|>$system \\
You are an intelligent urban mobility analysis model specializing in NYC taxi patterns. Your task is to predict the precise time interval until the next taxi event in Manhattan. You will be given a history of taxi trips, where each event includes its type (e.g. Uptown Pickup, Downtown Drop-off), the time elapsed since the last event, and a map image of the location. Analyze the patterns in event types and timing to forecast the time to the next event. Your response must be exactly four byte tokens representing the time interval. \\
$<|im\_end|>$\\
Event Sequence History:\\
$<|start\_of\_event|>$\\
$<|time\_start|>$$<|byte\_0|>$$<|byte\_0|>$$<|byte\_0|>$$<|byte\_0|>$$<|time\_end|>$\\
$<|type\_start|>$$<|type\_0|>$$<|type\_end|>$\\
$<|text\_start|>$Picked up at Tribeca (40.711086, -74.016106), 1 passengers.$<|text\_end|>$\\
$<|vision\_start|><|image\_pad|><|vision\_end|>$\\
$<|end\_of\_event|>$\\
$<|start\_of\_event|>$\\
$<|time\_start|>$$<|byte\_62|>$$<|byte\_162|>$$<|byte\_34|>$$<|byte\_34|>$$<|time\_end|>$\\
$<|type\_start|>$$<|type\_3|>$$<|type\_end|>$\\
$<|text\_start|>$Dropped off from Tribeca (40.711086, -74.016106) to Times Square (40.757698, -73.982124), 1 passengers, 2.87 miles trip.$<|text\_end|>$\\
$<|vision\_start|>$$<|image\_pad|>$$<|vision\_end|>$\\
$<|end\_of\_event|>$\\
$<|start\_of\_event|>$\\
$<|time\_start|>$$<|byte\_62|>$$<|byte\_196|>$$<|byte\_68|>$$<|byte\_68|>$$<|time\_end|>$\\
$<|type\_start|>$$<|type\_4|>$$<|type\_end|>$\\
$<|text\_start|>$Picked up at Upper West Side (40.799252, -73.970146), 1 passengers.$<|text\_end|>$\\
$<|vision\_start|>$$<|image\_pad|>$$<|vision\_end|>$\\
$<|end\_of\_event|>$\\
$<|time\_prediction|>$\\

\tcbline 

\textbf{Response:} \\
$<|byte\_63|>$$<|byte\_17|>$$<|byte\_16|>$$<|byte\_161|>$\\
\end{tcolorbox}

\begin{tcolorbox}[
  breakable,
  colback=black!5,
  colframe=black,
  arc=0mm,
  boxrule=0.5pt,
  left=5pt,
  right=5pt,
  top=5pt,
  bottom=5pt,
  width=\textwidth,
]
\captionof{table}{Prompt-response pair sample for next event type finetuning from TAXI-PRO}
\label{tab:s2-type}

\textbf{Prompt:} \\
$<|im\_start|>$system \\
You are an intelligent urban mobility analysis model specializing in NYC taxi patterns. Your task is to predict the type of the next taxi event in Manhattan based on a sequence of past events. You will be given a history of taxi trips, where each event includes its type (e.g. Uptown Pickup, Downtown Drop-off), the time elapsed since the last event, and a map image of the location. Analyze the patterns in timing and location to forecast the next action. Your response must be a single token representing one of the six event types. \\
$<|im\_end|>$\\
Event Sequence History:\\
$<|start\_of\_event|>$\\
$<|time\_start|>$$<|byte\_0|>$$<|byte\_0|>$$<|byte\_0|>$$<|byte\_0|>$$<|time\_end|>$\\
$<|type\_start|>$$<|type\_0|>$$<|type\_end|>$\\
$<|text\_start|>$Picked up at Tribeca (40.711086, -74.016106), 1 passengers.$<|text\_end|>$\\
$<|vision\_start|>$$<|image\_pad|>$$<|vision\_end|>$\\
$<|end\_of\_event|>$\\
$<|start\_of\_event|>$\\
$<|time\_start|>$$<|byte\_62|>$$<|byte\_162|>$$<|byte\_34|>$$<|byte\_34|>$$<|time\_end|>$\\
$<|type\_start|>$$<|type\_3|>$$<|type\_end|>$\\
$<|text\_start|>$Dropped off from Tribeca (40.711086, -74.016106) to Times Square (40.757698, -73.982124), 1 passengers, 2.87 miles trip.$<|text\_end|>$\\
$<|vision\_start|>$$<|image\_pad|>$$<|vision\_end|>$\\
$<|end\_of\_event|>$\\
$<|start\_of\_event|>$\\
$<|time\_start|>$$<|byte\_62|>$$<|byte\_196|>$$<|byte\_68|>$$<|byte\_68|>$$<|time\_end|>$\\
$<|type\_start|>$$<|type\_4|>$$<|type\_end|>$\\
$<|text\_start|>$Picked up at Upper West Side (40.799252, -73.970146), 1 passengers.$<|text\_end|>$\\
$<|vision\_start|>$$<|image\_pad|>$$<|vision\_end|>$\\
$<|end\_of\_event|>$\\
$<|type\_prediction|>$\\

\tcbline 

\textbf{Response:} \\
$<|type\_1|>$\\
\end{tcolorbox}

\section{The Use of Large Language Models}
In the preparation of this manuscript, we utilized LLM as a writing-assistive tool. Its application was strictly limited to improving the language and presentation of the paper, which included correcting grammatical errors, rephrasing sentences for enhanced clarity and conciseness, and refining paragraph structure. The core scientific contributions, including research ideation, methodology, and analysis, were conceived and executed entirely by the human authors. The authors have reviewed and edited all text and take full responsibility for the final content of the paper.

\end{document}